\documentclass[10pt,twocolumn,letterpaper]{article} 
\usepackage[pagenumbers]{theme} 
\renewcommand{\paragraph}[1]{\vspace{.5em}\noindent\textbf{#1.}}

\usepackage{subcaption}
\usepackage{booktabs}
\usepackage{multirow}
\usepackage[table]{xcolor}
\definecolor{goldhighlight}{RGB}{255, 235, 170}
\definecolor{silverhighlight}{RGB}{230, 230, 230} 
\definecolor{cvprblue}{rgb}{0.21,0.49,0.74} \usepackage[pagebackref,breaklinks,colorlinks,allcolors=cvprblue]{hyperref}

\title{MatteViT: High-Frequency-Aware Document Shadow Removal with Shadow
Matte Guidance}

\author{
Chaewon Kim\textsuperscript{*} \quad
Seoyeon Lee\textsuperscript{*} \quad
Jonghyuk Park\textsuperscript{$\dagger$}\\
Kookmin University\\
{\tt\small \{kimcwbf, tjdus0223, jonghyuk\}@kookmin.ac.kr}
}

\begin{document}
\maketitle

\begingroup
\renewcommand\thefootnote{}
\footnotetext{* Equal contribution.}
\footnotetext{$\dagger$ Corresponding author.}
\endgroup

\begin{abstract}
Document shadow removal is essential for enhancing the clarity of digitized documents. Preserving high-frequency details (e.g., text edges and lines) is critical in this process because shadows often obscure or distort fine structures. This paper proposes a matte vision transformer (MatteViT), a novel shadow removal framework that applies spatial and frequency-domain information to eliminate shadows while preserving fine-grained structural details. To effectively retain these details, we employ two preservation strategies. First, our method introduces a lightweight high-frequency amplification module (HFAM) that decomposes and adaptively amplifies high-frequency components. Second, we present a continuous luminance-based shadow matte, generated using a custom-built matte dataset and shadow matte generator, which provides precise spatial guidance from the earliest processing stage. These strategies enable the model to accurately identify fine-grained regions and restore them with high fidelity. Extensive experiments on public benchmarks (RDD and Kligler) demonstrate that MatteViT achieves state-of-the-art performance, providing a robust and practical solution for real-world document shadow removal. Furthermore, the proposed method better preserves text-level details in downstream tasks, such as optical character recognition, improving recognition performance over prior methods.
\end{abstract}    
\section{Introduction}
\label{sec:intro}

\begin{figure}[t]
    \centering
    
    \begin{subfigure}[b]{0.31\columnwidth}
        \includegraphics[width=\textwidth]{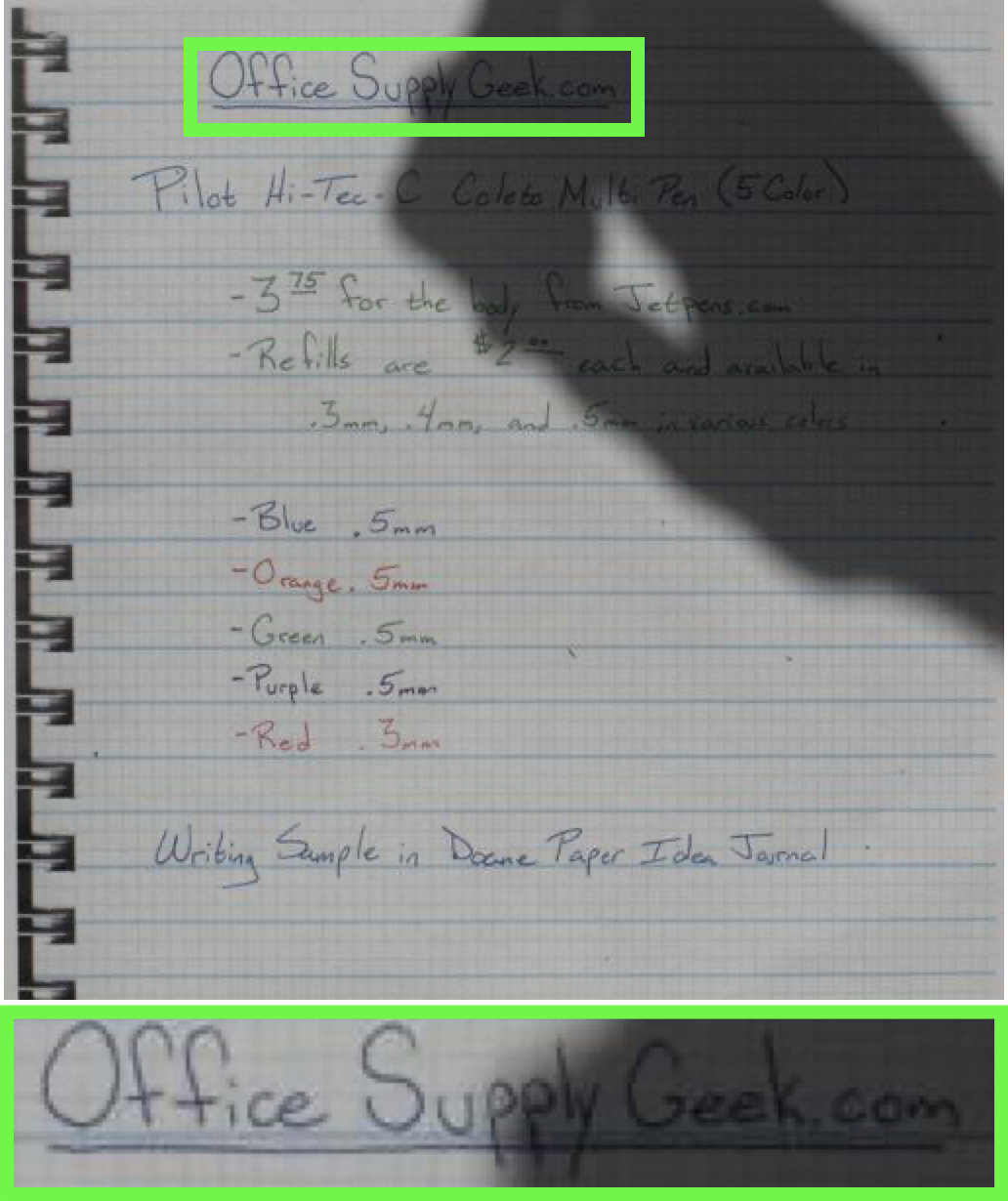}
        \caption{Input}
        \label{fig:intro_input}
    \end{subfigure}
    \hfill
    \begin{subfigure}[b]{0.31\columnwidth}
        \includegraphics[width=\textwidth]{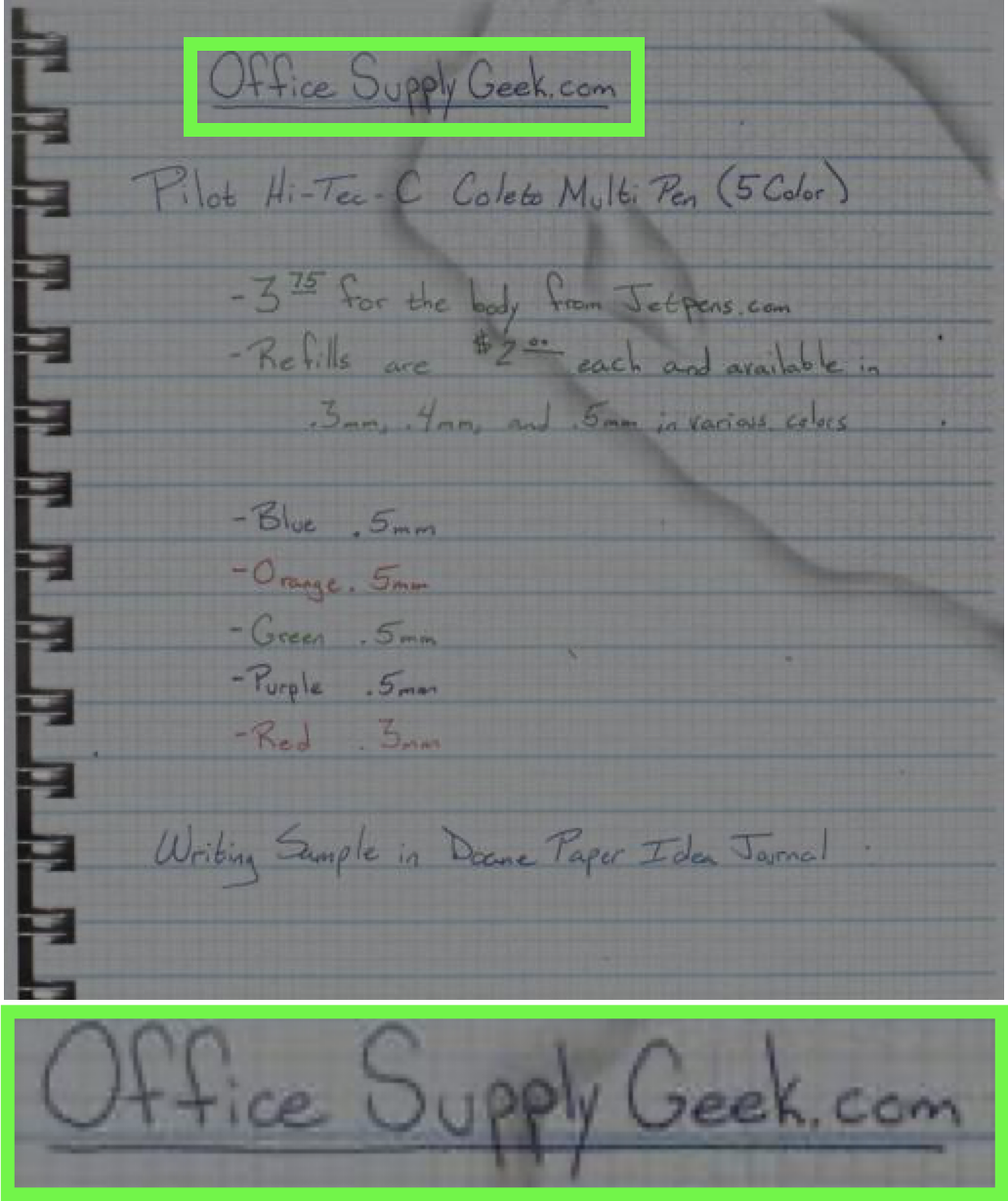}
        \caption{Wang et al.}
        \label{fig:intro_wang}
    \end{subfigure}
    \hfill
    \begin{subfigure}[b]{0.31\columnwidth}
        \includegraphics[width=\textwidth]{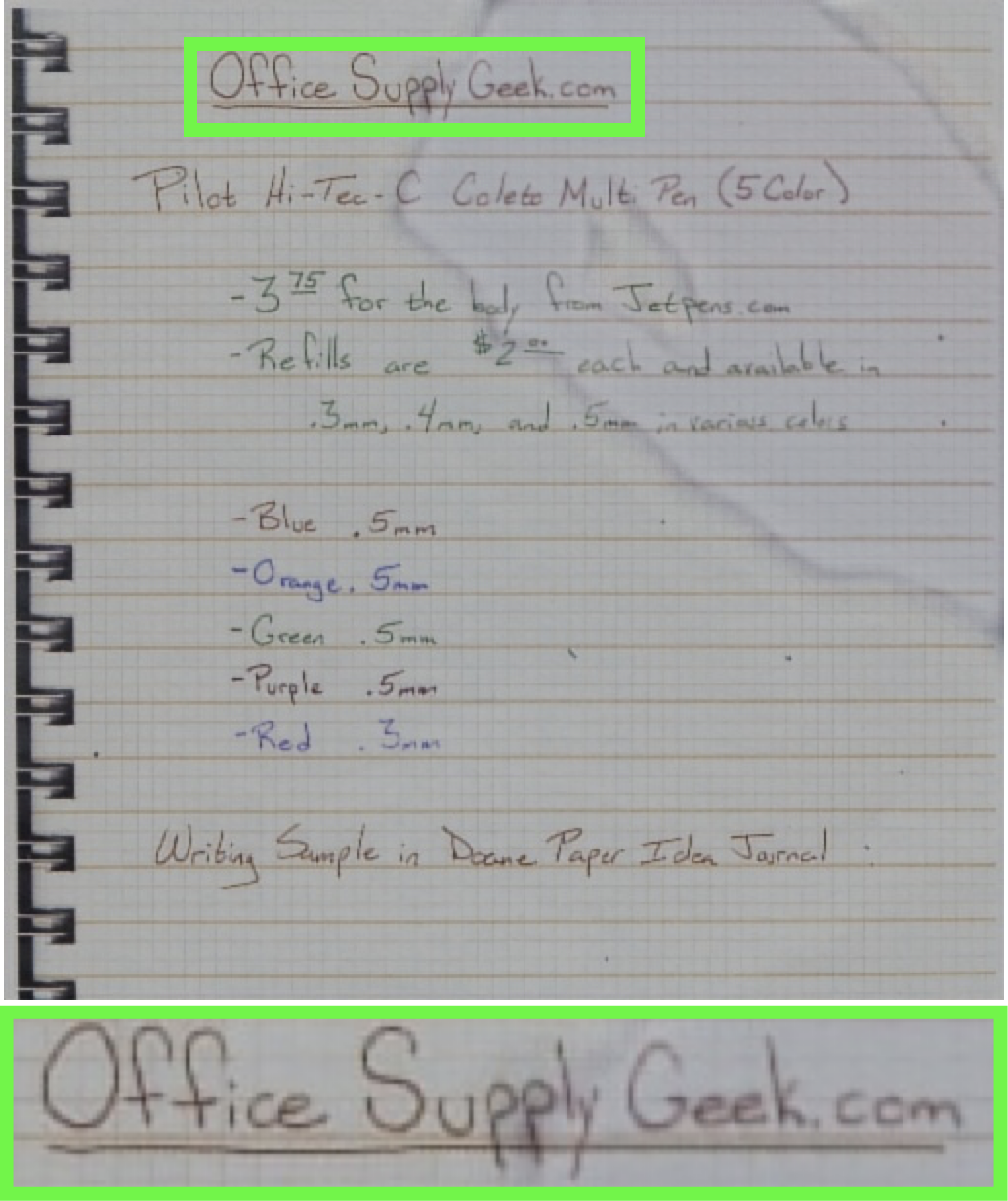}
        \caption{BEDSR-Net}
        \label{fig:intro_bedsrnet}
    \end{subfigure}
    
    \vspace{1mm}

    \begin{subfigure}[b]{0.31\columnwidth}
        \includegraphics[width=\textwidth]{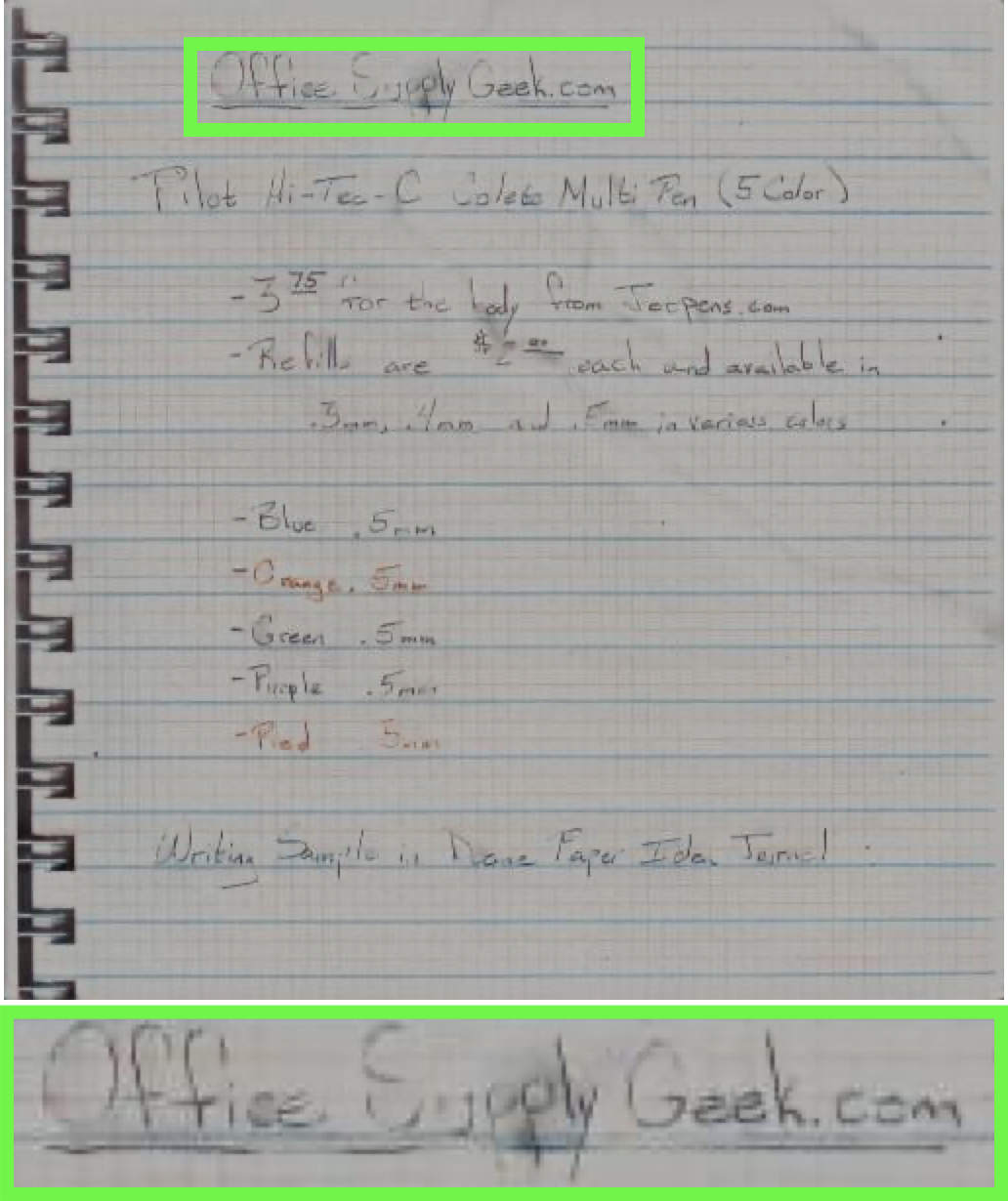}
        \caption{DocDeshadower}
        \label{fig:intro_dds}
    \end{subfigure}
    \hfill
    \begin{subfigure}[b]{0.31\columnwidth}
        \includegraphics[width=\textwidth]{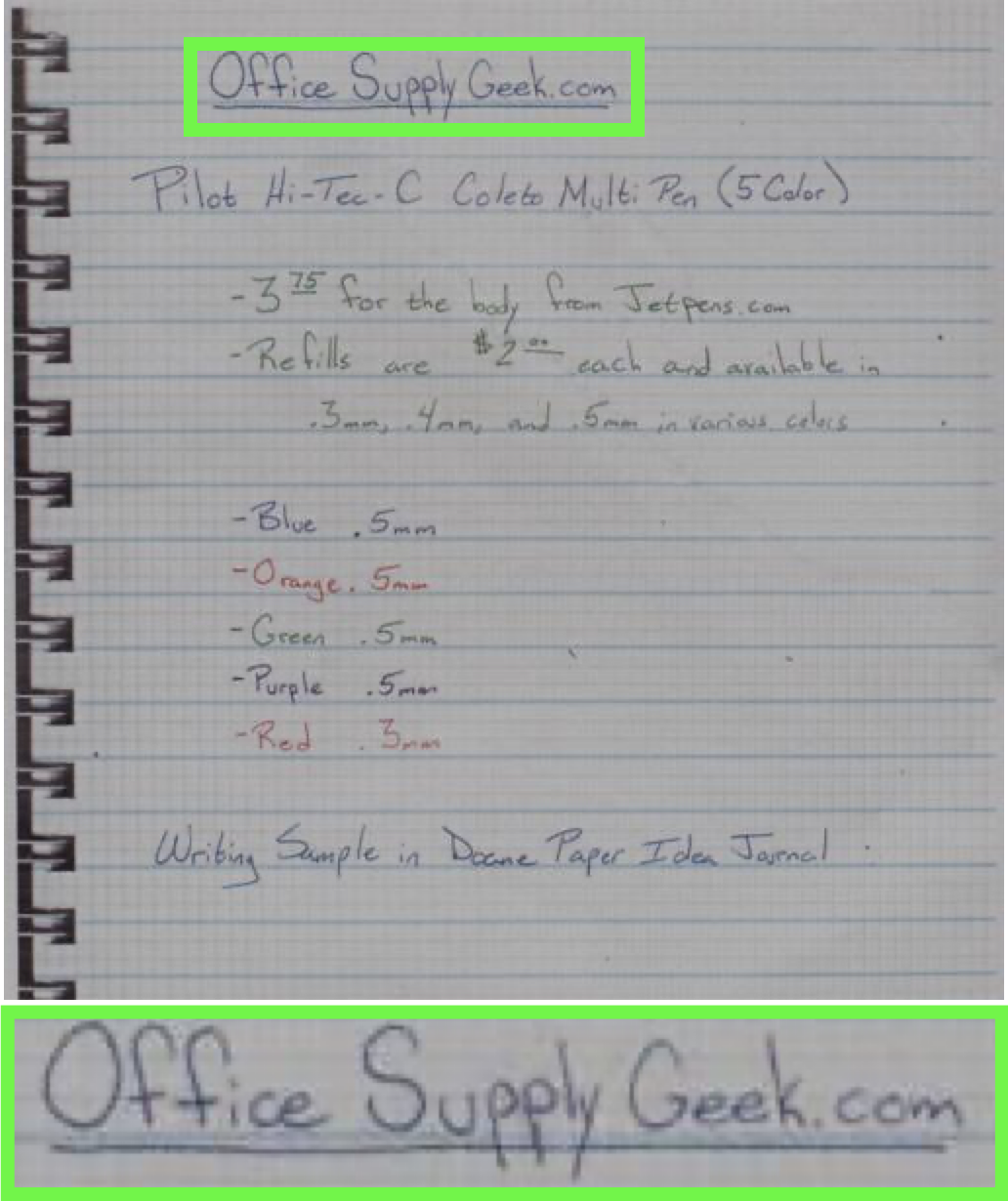}
        \caption{Ours}
        \label{fig:intro_ours}
    \end{subfigure}
    \hfill
    \begin{subfigure}[b]{0.31\columnwidth}
        \includegraphics[width=\textwidth]{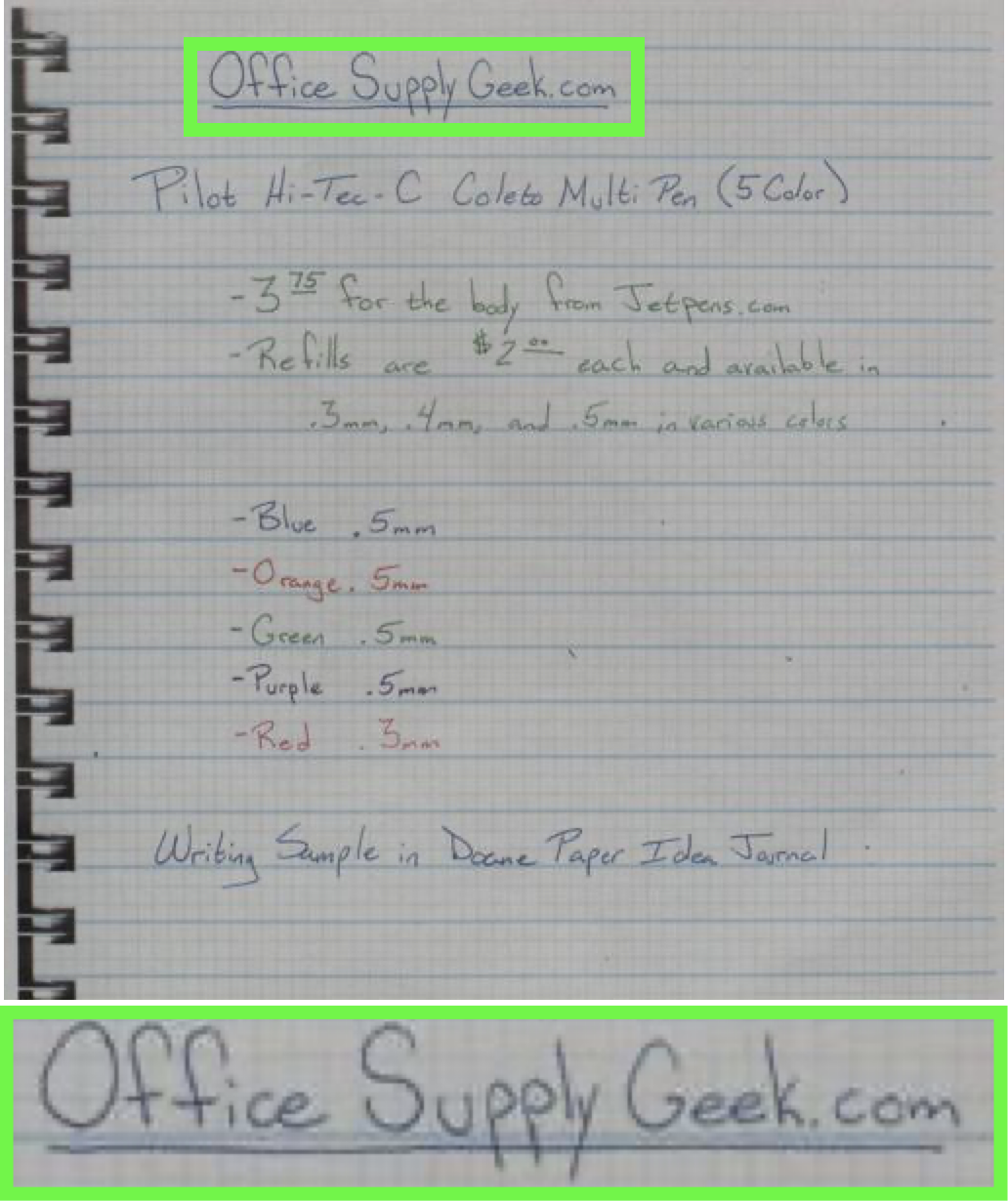}
        \caption{Target}
        \label{fig:intro_target}
    \end{subfigure}
    
    \caption{Qualitative comparison of shadow removal methods on document images.}
    \label{fig:intro_comparison}
\end{figure}

The increasing demand for automated document digitization and analysis in diverse domains, such as finance, education, health care,  and public services, has significantly increased the importance of high-quality images of digital documents~\cite{li2023high, wang2025comprehensive}. These digitized documents support a broad range of downstream applications, including information extraction, intelligent archiving, and digital accessibility~\cite{zhang2023document}. However, real-world document acquisition regularly occurs under uncontrolled conditions, resulting in visual degradation, including blurring, uneven illumination, and shadows~\cite{wang2022udoc, Anvari2021arXiv, chen2023shadocnet}.

Among these problems, shadows caused by uneven lighting or physical obstructions remain a persistent and challenging form of degradation~\cite{pei2023doc, liu2025leveraging, liu2021shadow}. Shadows obscure visual content and hinder the preservation of fine-grained structural details (e.g., text strokes and layout elements)~\cite{feng2021doctr}. Therefore, shadows significantly degrade readability and the performance of downstream tasks, including optical character recognition (OCR), layout analysis, and information retrieval~\cite{souibgui2022docentr, lu2017shadow}. Even subtle or soft shadows can interrupt the continuity of textual and structural elements~\cite{georgiadis2023lp}. Therefore, effective shadow removal has become an imperative preprocessing step in document processing. 

Despite recent progress, removing shadows while preserving the fine structural details of documents remains challenging~\cite{zhou2024docdeshadower, chen2024shadocformer}. Various elements (e.g., character strokes, table lines, edge contours, and background textures) contain essential semantic and structural information. Preserving high-frequency details is critical for human readability and the reliability of machine-based interpretation. Once degraded, these components can significantly hamper OCR performance and decrease the overall integrity of the document image~\cite{yang2023docdiff, fu2019cascaded}.

\Cref{fig:intro_comparison} reveals that many shadow removal methods fail to recover these structural details completely and are ineffective at removing shadows thoroughly. Although some methods can suppress shadows to a degree, they often smooth out or distort nearby content~\cite{chang2023tsrformer}. Coarse shadow representations fail to capture gradual luminance transitions and soft shadow boundaries, leading to incomplete removal of visual artifacts~\cite{le2020shadow, guo2023shadowdiffusion}. Approaches that ignore these fine structures tend to produce excessively smooth outputs without essential information.

This paper presents a matte vision transformer (MatteViT)  framework designed for structure-preserving document shadow removal to address these limitations. Our main contributions are summarized as follows:

\begin{itemize}
    \item To preserve the fine details often lost during shadow removal, we introduce the High-Frequency Amplification Module (HFAM), which selectively enhances high-frequency components while maintaining structural integrity.
    
    \item We construct new shadow matte datasets based on the RDD and Kligler benchmarks, providing soft luminance supervision for robust shadow matte learning. These datasets will be made publicly available upon acceptance. Leveraging them, we propose a novel guidance approach based on a continuous luminance-based shadow matte, generated by a shadow matte generator. This guidance captures shadow intensity and location for precise localization while preserving structural details.

    \item Extensive experiments on two benchmark datasets (RDD and Kligler) demonstrate that our method achieves state-of-the-art performance in document shadow removal. Furthermore, it shows superior OCR accuracy, validating its effectiveness in preserving text-level details essential for practical document digitization.

\end{itemize}

We demonstrated that MatteViT removes shadows while preserving fine-grained structure. By achieving state-of-the-art performance on benchmarks, it provides a robust solution for real-world document restoration, enhancing clarity and interpretability.
\section{Related Work}
\label{sec:related_work}

\subsection{Shadow Removal Without Mask Supervision}

Recent studies on enhancing practicality in real-world applications have increasingly focused on removing document shadows without employing pixel-level mask annotations. These mask-free approaches provide better scalability and adaptability by directly training models on shadowed and shadow-free image pairs.

For instance, DeShadowNet~\cite{qu2017deshadownet} employs a multicontext architecture integrating global and semantic features to predict shadow mattes with fine local details. The LG-ShadowNet~\cite{liu2021shadow} method employs a two-stage approach for unpaired shadow removal, where a lightness-guided convolutional neural network adjusts the illumination inconsistencies, and a network applies this guidance to remove shadows while preserving structural details. The single-stage decoupled multitask network (DMTN)~\cite{liu2023decoupled} explicitly decomposes features for joint learning of shadow removal, matte estimation, and shadow reconstruction. The three-branch residual network TBRNet~\cite{liu2023shadow} jointly models shadow removal, shadow matte estimation, and shadow image reconstruction in a single-stage framework. In addition, ShaDocFormer~\cite{chen2024shadocformer} employs a transformer-based architecture combining threshold-guided attention with cascaded refinement for accurate shadow detection and removal. The DocDeshadower~\cite{zhou2024docdeshadower} approach employs a multifrequency transformer architecture based on the Laplacian pyramid to remove shadows while preserving the document layout and texture.

Despite their strong generalizability, the absence of explicit spatial supervision makes it challenging to localize shadows accurately, especially those with soft edges or gradual intensity changes.

\subsection{Shadow Removal With Mask Supervision}

Other approaches have incorporated explicit shadow masks to provide spatial cues during training or inference. These masks guide the model in identifying shadow regions, allowing more precise correction.

For instance, the mask-shadow generative adversarial network (Mask-ShadowGAN)~\cite{hu2019mask} jointly learns shadow mask prediction, generation, and removal via adversarial training with cycle-consistency constraints. The document-specific background estimation document shadow removal network (BEDSR-Net)~\cite{lin2020bedsr} method separates background estimation and shadow elimination to improve readability and visual quality. The ShadowFormer~\cite{guo2023shadowformer} approach applies a retinex-based transformer with shadow-interaction attention to model the global context between shadowed and nonshadowed regions.

\begin{figure*}[t]
    \centering
    \includegraphics[width=0.9\textwidth]{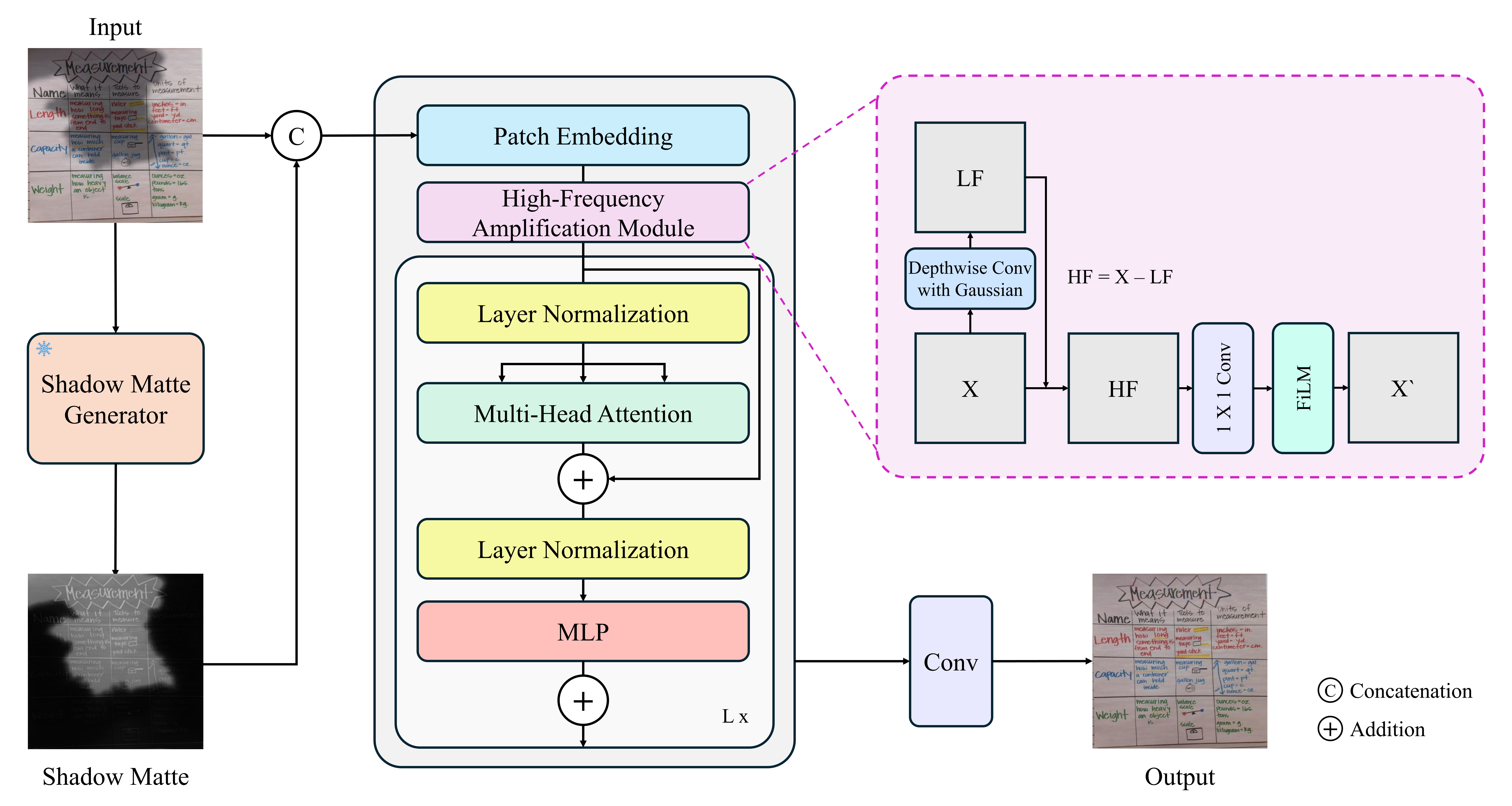}
    \caption{Overview of the proposed matte vision transformer (MatteViT) framework. Given an input shadow image, a shadow matte is predicted and concatenated with the input. The combined representation is input into a ViT-based network equipped with a high-frequency amplification module (HFAM) to recover the shadow-free image.}
    \label{fig:framework}
\end{figure*}

Despite their effectiveness, these methods depend on manually annotated binary masks, which struggle to capture soft transitions and continuous shadow intensities regularly found in real-world scenarios.

Thus, this paper proposes MatteViT, a framework that applies a continuous shadow matte framework for fine-grained spatial guidance to overcome these limitations. Unlike binary masks, the matte guidance encodes the shadow location and intensity, enabling soft, structure-aware conditioning. This approach permits the model to retain high-frequency details that are essential for readability without requiring external annotations.
\section{Method}
\label{sec:method}

This work proposes MatteViT,  a novel framework for document shadow removal that applies shadow matte information and frequency-aware enhancement techniques. The proposed method integrates a shadow matte to represent shadow regions, enhances high-frequency details via an HFAM, and applies frequency-sensitive loss functions. The self-attention mechanism in the ViT~\cite{dosovitskiy2020image} allows the proposed model to focus more effectively on shadow-affected regions while preserving the overall document structure. ~\cref{fig:framework} presents the overall architecture, and Sections 3.1 to 3.3 detail each component.

\subsection{Shadow Matte Generator}

This approach generates a shadow matte representing the intensity of shadow regions in a continuous manner to guide the shadow removal process. Rather than relying on conventional binary shadow masks, this matte captures fine luminance variations between shadowed and nonshadowed regions, offering richer and more detailed spatial guidance. Unlike binary masks, the proposed matte preserves soft transitions and structural details, helping the network more precisely localize and interpret shadow boundaries.

\begin{figure}[h]
    \centering
    \includegraphics[width=\columnwidth]{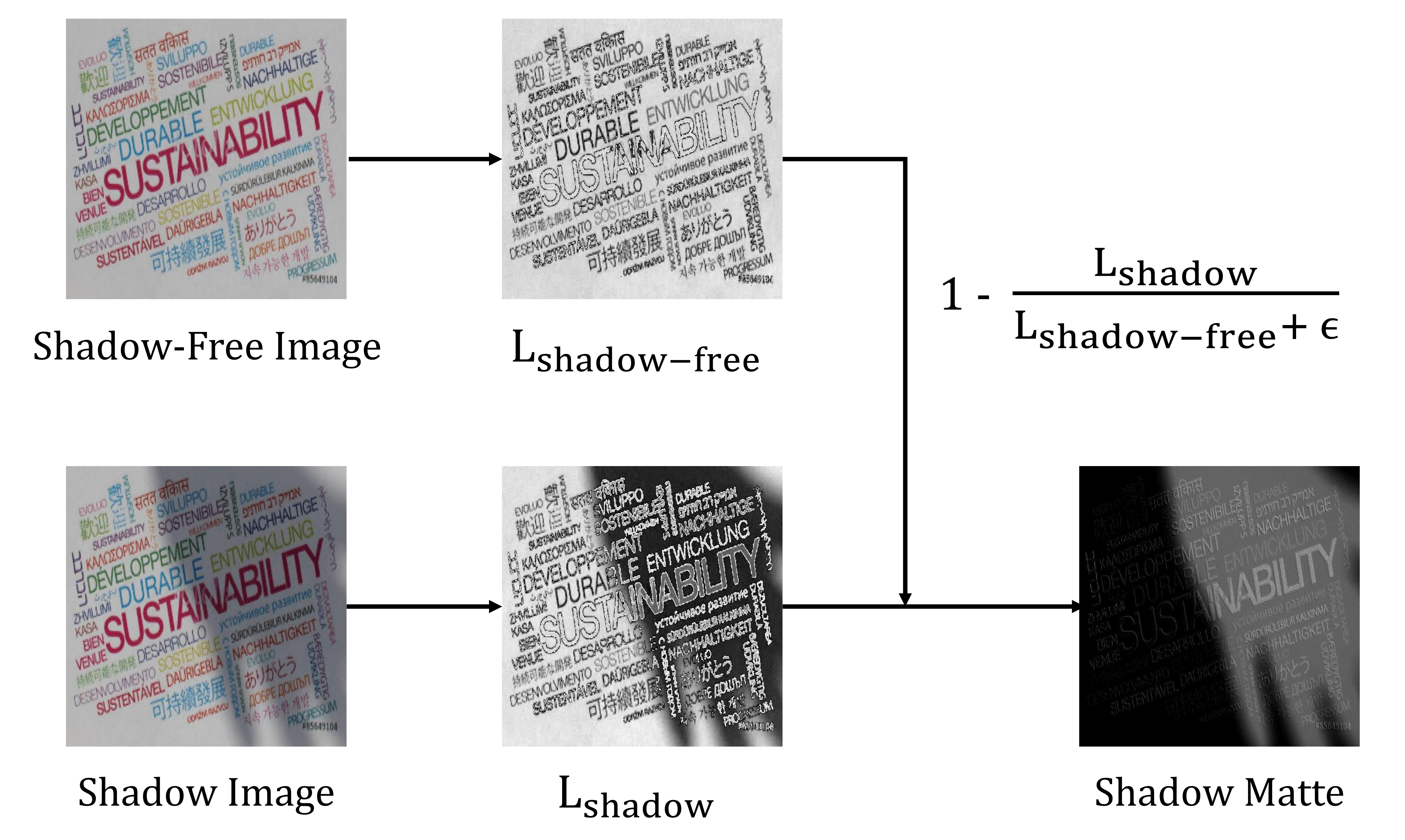}
    \caption{Shadow matte generation by computing the luminance difference between paired shadow and shadow-free images. The shadow image serves as input,  and the computed matte is the target for training the shadow matte generator.}
    \label{fig:matte_process}
\end{figure}

This method constructs a dataset of shadow mattes computed from paired shadow and shadow-free images, as illustrated in ~\cref{fig:matte_process}, to train a network capable of predicting the shadow matte results. Both images are converted to the LAB color space, and the luminance (L) channels of the shadow and shadow-free images, denoted as $L_{\text{shadow}}$ and $L_{\text{shadow-free}}$, are extracted. Then, the matte is calculated as the normalized difference between these luminance values:

\begin{equation}
\text{Shadow Matte} = 1 - \frac{L_{\text{shadow}}}{L_{\text{shadow-free}} + \epsilon},
\label{eq:shadow_matte}
\end{equation}

where $\epsilon$ denotes a small constant to prevent division by zero. The resulting matte has values ranging between 0 and 1, highlighting shadowed areas with higher values and smoothly transitioning to zero in well-lit regions.

The shadow matte generator is trained using the U-Net~\cite{ronneberger2015u} architecture, with the shadow image as input and the computed shadow matte value as the ground truth. To optimize the generator, we employ a composite loss function combining L1 loss and a binary cross-entropy loss, balanced with weights of 0.7 and 0.3, respectively. This combination effectively balances accurate pixel-level intensity regression (L1) with stable probabilistic classification (from the cross-entropy loss), leading to more precise matte predictions. Once trained, this network is frozen and applied as a fixed preprocessing module for shadow removal.

\cref{fig:framework} reveals that MatteViT explicitly concatenates the predicted shadow matte with the input image and inputs it into a ViT-based network, providing early spatial and intensity cues that focus the model on shadow-affected regions.

\subsection{High-Frequency Amplification Module}

Document shadow removal requires the careful preservation and enhancement of fine-grained details (e.g., text edges, line strokes, and document textures). These high-frequency components are often degraded or lost during the shadow removal process. This work introduces the HFAM to address this challenge, which selectively enhances high-frequency information while maintaining the overall structural integrity of the document.

The HFAM operates immediately after patch embedding in the ViT architecture. Given the feature representation from the patch embedding layer, HFAM first decomposes the features into low- and high-frequency components via a frequency separation process.

\paragraph{Frequency Decomposition} A depthwise convolution with a 5×5 Gaussian kernel extracts low-frequency components, with the kernel initialized using a normalized 2D Gaussian distribution. The high-frequency component is obtained by subtracting the low-frequency features from the original input.

\paragraph{Adaptive Amplification} We designed an adaptive modulation mechanism inspired by feature to amplify high-frequency details selectively~\cite{perez2018film}. High-frequency features are processed through global average pooling and a lightweight multilayer perceptron to generate channelwise scaling and shifting parameters. The enhanced features are calculated as follows:

\begin{equation}
\mathbf{X}' = \mathbf{X} + \gamma \odot \mathbf{X}_{\text{HF}} + \beta,
\label{eq:feature_enhancement}
\end{equation}

$\text{X}$ represents the original features, $\text{X}_{\text{HF}}$ denotes a high-frequency component, $\gamma$ and $\beta$ indicate the learned scaling and shifting parameters, respectively, and $\odot$ represents element-wise multiplication.

This design allows the network to adjust the level of high-frequency enhancement dynamically according to the input content, more precisely emphasizing the fine structures (e.g., text edges and subtle textures) that are critical in document images. This approach adds minimal computational overhead while considerably improving the preservation of detailed visual information.

\subsection{Loss Functions}

The proposed approach for practical model training involves a composite loss function addressing spatial reconstruction quality and frequency-domain characteristics. This approach combines the weighted Charbonnier loss for spatial fidelity with the fast Fourier transform (FFT)  loss to encourage frequency-aware reconstruction.

\paragraph{Edge-aware Charbonnier Loss} As the primary reconstruction objective, this approach adopts the Charbonnier loss~\cite{charbonnier1997deterministic} due to its robustness to outliers and stable gradient behavior, making it suitable for fine-grained image restoration. We extend this loss with an adaptive weighting scheme based on edge information to emphasize high-frequency regions. Specifically, the weight map is computed using a Laplacian operator applied to the ground-truth image, highlighting the edges and textures:

\begin{equation}
\mathcal{L}_{\text{char}} = \mathbb{E}\left[\sqrt{w \cdot (I_{\text{pred}} - I_{\text{gt}})^2 + \epsilon^2}\right],
\label{eq:char_loss}
\end{equation}

where $w$ denotes the high-frequency weight derived from the Laplacian response, $I_{pred}$ and $I_{gt}$ represent  the predicted and ground-truth images, respectively, and $\epsilon$ indicates a small constant for numerical stability. This formulation encourages the network to preserve fine details and sharp transitions that are critical for document quality.

\paragraph{FFT Loss} The FFT-based loss minimizes the discrepancy between the frequency components of the predicted and ground-truth images to ensure spectral consistency:
 \begin{equation}
\mathcal{L}_{\text{fft}} = \mathbb{E}\left[|\mathcal{F}(I_{\text{pred}}) - \mathcal{F}(I_{\text{gt}})|\right],
\label{eq:fft_loss}
\end{equation}
where $\mathcal{F}$ denotes the 2D FFT, which helps maintain global structures and local textures in the frequency domain.

\begin{table*}[t]
  \centering
  \caption{Quantitative results of comparisons with state-of-the-art methods on the RDD and Kligler datasets. The best results are in bold with a gold background, and the second best are on a silver background.}
  \label{tab:main_comparison}
  
  \begin{tabular}{@{}l ccc ccc@{}}
    \toprule
    \multirow{2}{*}{Method} 
    & \multicolumn{3}{c}{RDD (512$\times$512)} 
    & \multicolumn{3}{c}{Kligler (512$\times$512)} \\
    \cmidrule(lr){2-4} \cmidrule(lr){5-7}
    & PSNR $\uparrow$ & SSIM $\uparrow$ & RMSE $\downarrow$ 
    & PSNR $\uparrow$ & SSIM $\uparrow$ & RMSE $\downarrow$ \\
    \midrule
    Shah \textit{et al.} (2018)~\cite{shah2018iterative} & 10.51 & 0.79 & 81.72 & 8.91 & 0.78 & 91.62 \\
    Jung \textit{et al.} (2018)~\cite{jung2018water} & 15.16 & 0.82 & 48.75 & 14.69 & 0.84 & 47.55 \\
    Wang \textit{et al.} (2019)~\cite{wang2019effective} & 15.22 & 0.8 & 48.04 & 16.47 & 0.83 & 41.64 \\
    Wang \textit{et al.} (2020)~\cite{wang2020shadow} & 15.82 & 0.82 & 46.81 & 15.55 & 0.82 & 45.08 \\
    Liu \textit{et al.} (2023)~\cite{liu2023shadow} & 20.47 & 0.86 & 26.39 & 21.67 & 0.83 & 21.93 \\
    DeShadowNet (2017)~\cite{qu2017deshadownet} & 18.48 & 0.86 & 32.31 & 23.47 & 0.82 & 17.98 \\
    ST-CGAN (2020)~\cite{qi2020stc} & 17.95 & 0.61 & 32.07 & 16.57 & 0.75 & 37.16 \\
    BEDSR-Net (2020)~\cite{lin2020bedsr} & 25.32 & 0.85 & 14.53 & 22.88 & 0.74 & 21.02 \\
    LG-ShadowNet (2021)~\cite{liu2021shadow} & \cellcolor{silverhighlight}29.78 & 0.93 & 9.85 & 26.95 & \cellcolor{silverhighlight}0.92 & 13.88 \\
    DMTN (2023)~\cite{liu2023decoupled} & 25.84 & 0.94 & 13.90 & 27.11 & 0.89 & \cellcolor{silverhighlight}12.42 \\
    ShaDocFormer (2024)~\cite{chen2024shadocformer} & 29.4 & \cellcolor{silverhighlight}0.96 & \cellcolor{silverhighlight}9.16 & 25.65 & 0.9 & 15.66 \\
    DocDeShadower (2024)~\cite{zhou2024docdeshadower} & 28.78 & \cellcolor{silverhighlight}0.96 & 9.93 & \cellcolor{silverhighlight}27.17 & 0.91 & 12.61 \\
    \midrule
    Ours & \cellcolor{goldhighlight}\textbf{33.78} & \cellcolor{goldhighlight}\textbf{0.97} & \cellcolor{goldhighlight}\textbf{5.76} & \cellcolor{goldhighlight}\textbf{29.2} & \cellcolor{goldhighlight}\textbf{0.94} & \cellcolor{goldhighlight}\textbf{11.43} \\
    \bottomrule
  \end{tabular}
\end{table*}

\paragraph{Total Loss} The final training objective is defined as a weighted combination of both loss components:
\begin{equation}
\mathcal{L}_{\text{total}} = \mathcal{L}_{\text{char}} + \lambda \cdot \mathcal{L}_{\text{fft}},
\label{eq:total_loss}
\end{equation}
where $\lambda$ balances the contribution of the frequency loss. We conducted experiments by varying the hyperparameter $\lambda$ over the values $\{0.01, 0.05, 0.1, 0.5\}$, and selected $\lambda = 0.1$ based on empirical performance.

In summary, the MatteViT framework integrates shadow matte guidance, high-frequency amplification, and frequency-sensitive loss functions to achieve precise and visually consistent document shadow removal. The comprehensive experiments presented in the following sections demonstrate the effectiveness of this approach.
\section{Experiments}
\label{sec:experiments}

\subsection{Experimental Setup}

\paragraph{Datasets} The experiments were conducted on two publicly available document shadow removal datasets: RDD~\cite{zhang2023document} and Kligler~\cite{kligler2018document}, both providing paired shadow and shadow-free images for supervised training. In addition, we constructed shadow matte datasets for each of the two benchmarks to provide soft luminance-based supervision. The datasets will be made publicly available upon acceptance. All images were resized to 512×512 resolution during training and evaluation.

\paragraph{Implementation Details} The shadow matte generator was trained for 200 epochs using an RMSprop optimizer with a learning rate of 5e-6 and a batch size of 8. The MatteViT network was trained using the Adam optimizer at a learning rate of 0.0004. Its patch embedding employs a grid size of 4x4, which was empirically determined according to the input image size. All experiments were conducted on a single NVIDIA RTX 4090 GPU.

\paragraph{Evaluation Metrics} The quantitative evaluation applies three standard image quality metrics: peak signal-to-noise ratio (PSNR), structural similarity index measure (SSIM), and root mean square error (RMSE). These metrics assess the reconstruction fidelity from pixelwise and perceptual perspectives.

\paragraph{Baselines} The proposed method was evaluated in comparison with state-of-the-art shadow removal approaches. The baselines include traditional methods and deep learning-based models.

\begin{figure*}[t]
  \centering
  
  \begin{subfigure}[b]{0.135\textwidth}
      \includegraphics[width=\textwidth]{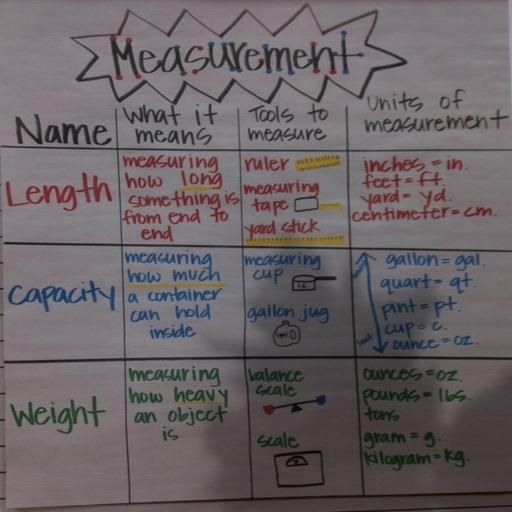}
      \vspace{1mm}
      \includegraphics[width=\textwidth]{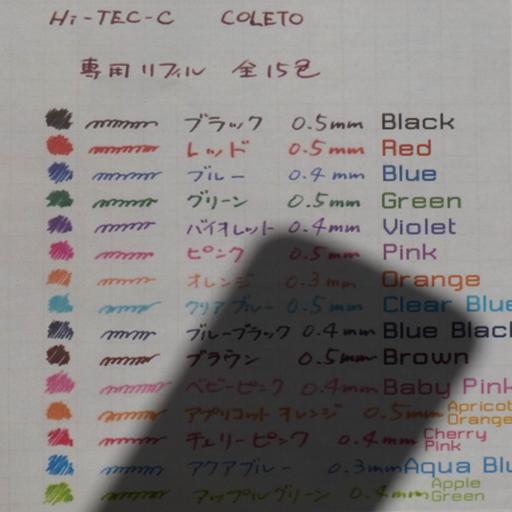}
      \caption{Input}
      \label{fig:qual_input}
  \end{subfigure}
  \hfill
  \begin{subfigure}[b]{0.135\textwidth}
      \includegraphics[width=\textwidth]{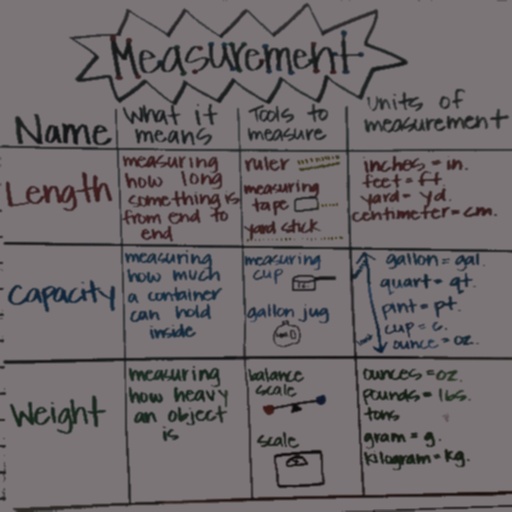}
      \vspace{1mm}
      \includegraphics[width=\textwidth]{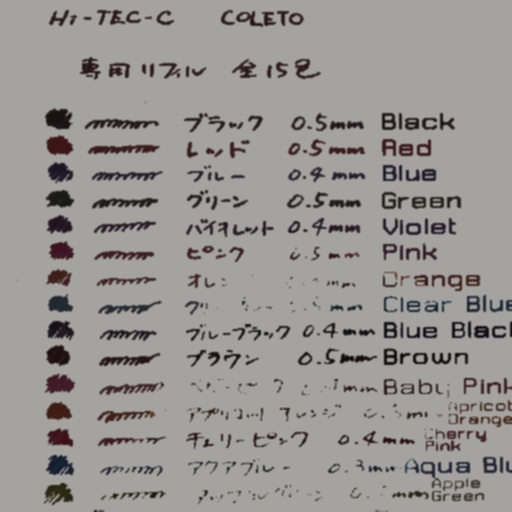}
      \caption{Liu et al.}
      \label{fig:qual_liu}
  \end{subfigure}
  \hfill
  \begin{subfigure}[b]{0.135\textwidth}
      \includegraphics[width=\textwidth]{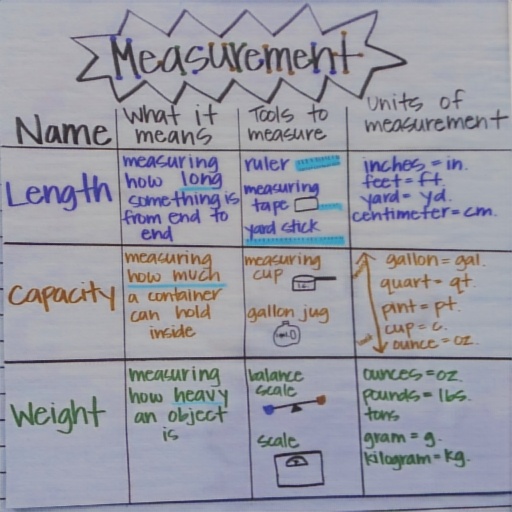}
      \vspace{1mm}
      \includegraphics[width=\textwidth]{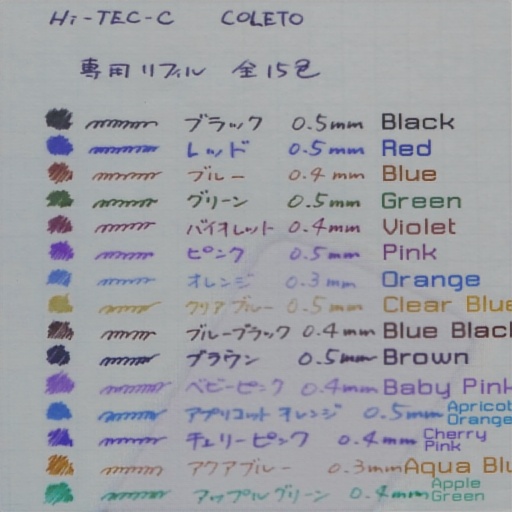}
      \caption{BEDSR-Net}
      \label{fig:qual_bedsr}
  \end{subfigure}
  \hfill
  \begin{subfigure}[b]{0.135\textwidth}
      \includegraphics[width=\textwidth]{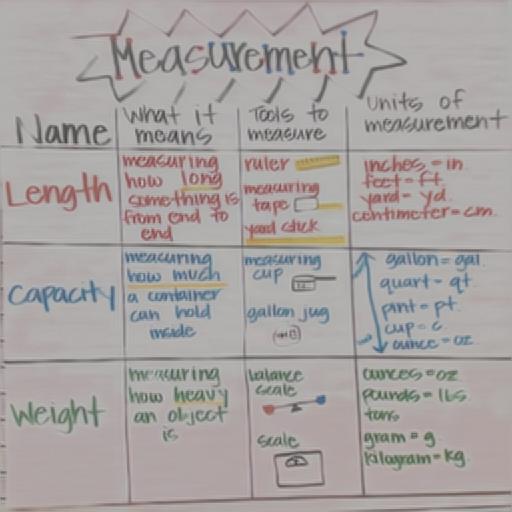}
      \vspace{1mm}
      \includegraphics[width=\textwidth]{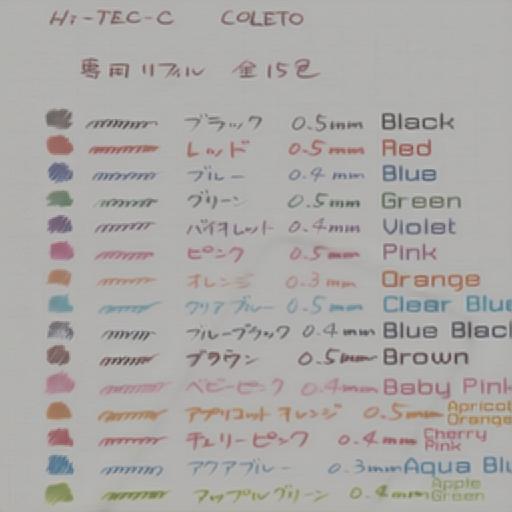}
      \caption{DeShadowNet}
      \label{fig:qual_deshadownet}
  \end{subfigure}
  \hfill
  \begin{subfigure}[b]{0.135\textwidth}
      \includegraphics[width=\textwidth]{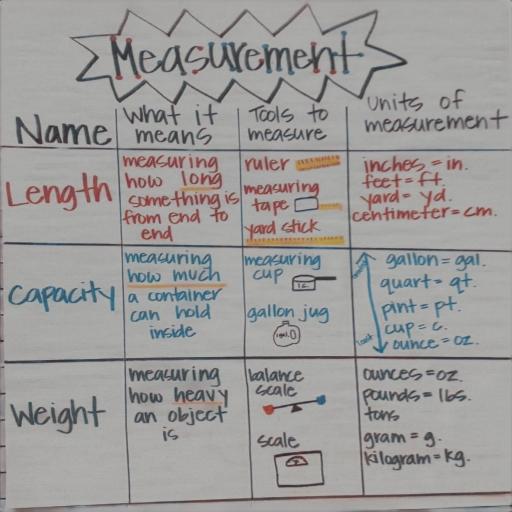}
      \vspace{1mm}
      \includegraphics[width=\textwidth]{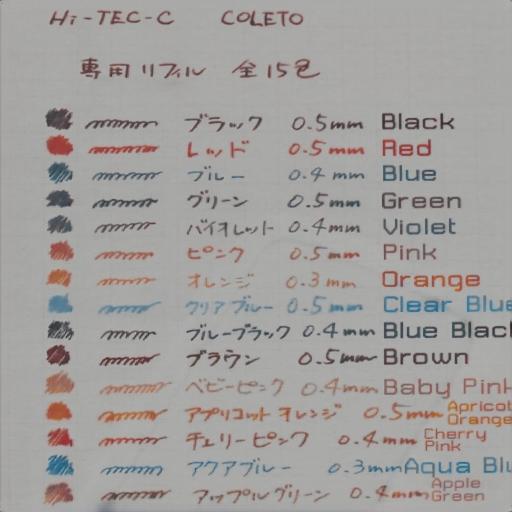}
      \caption{DocDeShadower}
      \label{fig:qual_dds}
  \end{subfigure}
  \hfill
  \begin{subfigure}[b]{0.135\textwidth}
      \includegraphics[width=\textwidth]{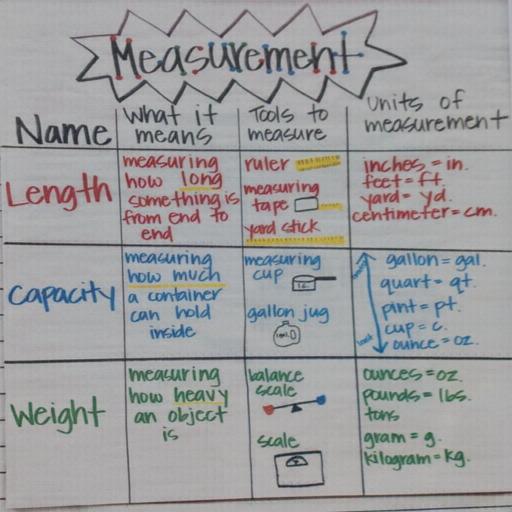}
      \vspace{1mm}
      \includegraphics[width=\textwidth]{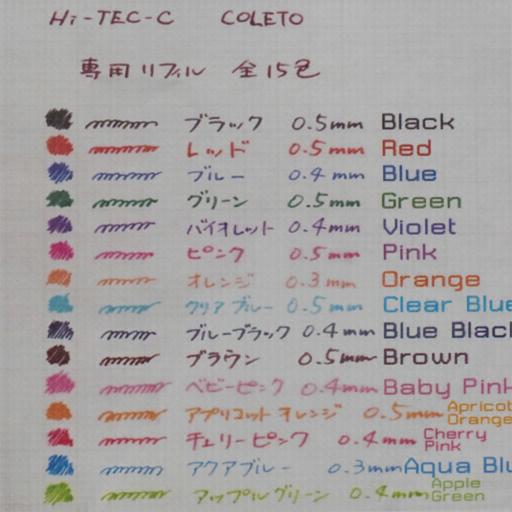}
      \caption{Ours}
      \label{fig:qual_ours}
  \end{subfigure}
  \hfill
  \begin{subfigure}[b]{0.135\textwidth}
      \includegraphics[width=\textwidth]{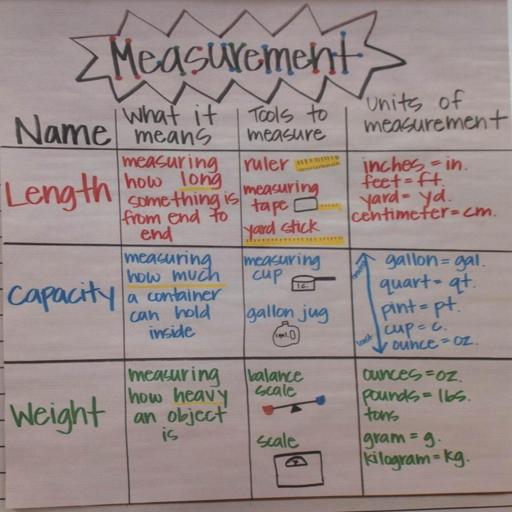}
      \vspace{1mm}
      \includegraphics[width=\textwidth]{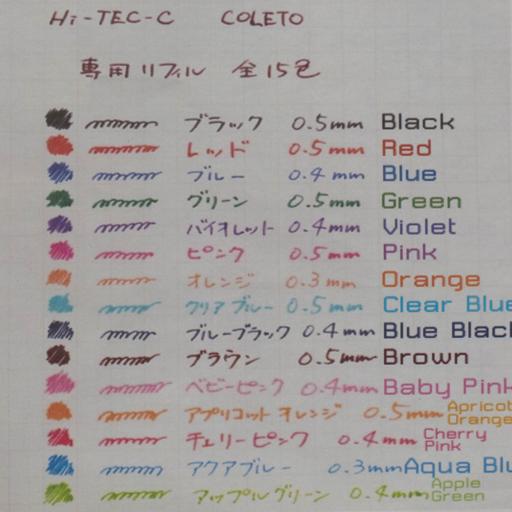}
      \caption{Target}
      \label{fig:qual_target}
  \end{subfigure}

  \caption{Qualitative comparison of shadow removal methods for two examples.}
  \label{fig:main_shadow_comparison}
\end{figure*}

\subsection{Comparison with State-of-the-Art Methods}

This work evaluates the performance of the proposed method against several state-of-the-art approaches on the RDD and Kligler datasets (\cref{tab:main_comparison}).

On the RDD dataset, the proposed method achieves a PSNR of 33.78  dB and the highest SSIM of 0.97, while significantly reducing the RMSE to 5.76. These results represent a notable improvement over the existing methods and set a new state-of-the-art level in document shadow removal, especially under complex shadow conditions.

On the Kligler dataset, MatteViT delivers the best performance, achieving a PSNR of 29.2 dB, SSIM of 0.94, and the lowest RMSE of 11.43. This outcome demonstrates the robustness and generalizability of this approach across diverse document types and illumination variations.

The proposed method removes shadows while preserving fine details, significantly improving performance.  By guiding the model with a continuous-valued shadow matte, the network can more precisely focus on shadowed regions and preserve subtle luminance transitions. Furthermore, the HFAM enhances local texture and edge information, which is beneficial for recovering fine text details that are frequently degraded by shadows. In addition, the frequency-sensitive loss functions help the model maintain the fine details and overall visual coherence.

In ~\cref{fig:main_shadow_comparison}, these strengths translate into clearer backgrounds and sharper textual content in the final output. The proposed method preserves high-frequency details (e.g., lines and text contours), maintains color information (e.g., text hues), and more thoroughly removes shadows than other approaches.

\subsection{OCR Performance Evaluation}

\begin{table}[b]
  \centering
  \caption{Optical character recognition (OCR) performance comparison measured by edit distance between the text recognition results for the ground-truth and restored images.}
  \label{tab:ocr_results}
  \begin{tabular}{@{}lc@{}}
    \toprule
    Method & Edit Distance $\downarrow$ \\
    \midrule
    Liu et al. & 142.43 \\
    BEDSR-Net & 125.07 \\
    ShaDocFormer & 149.79 \\
    DocDeShadower & 138.68 \\
    \midrule
    Ours & \textbf{113.21} \\
    \bottomrule
  \end{tabular}
\end{table}

We evaluate the OCR performance on the processed images to assess the practical utility of the shadow removal method for document processing applications. Removing shadows often improves the overall image clarity; however, removal can sometimes result in the loss of fine text details. Therefore, this evaluation measures how well the proposed method preserves text readability, a critical requirement for real-world document digitization tasks. For this purpose, we compared the proposed approach with several document shadow removal models, applying OCR using Tesseract~\cite{smith2007overview} to the shadow-free ground-truth images and the restored output from each method.

Then, the edit distance between the extracted text results is measured. The edit distance measures the minimum number of character-level operations (insertions, deletions, or substitutions) required to transform one text sequence into another.

\cref{tab:ocr_results} indicates that MatteViT performed best compared with the baseline methods. This improvement is primarily attributed to the HFAM (which enhances fine edges and text contours that are often degraded by shadows) and the frequency-sensitive loss functions that help preserve local details and global consistency during restoration. By preventing over-smoothing and retaining sharp character boundaries, the proposed method ensures that the textual content remains highly legible, generating more accurate OCR results.

\subsection{Effectiveness of the Shadow Matte Generator}

\begin{table}[b]
  \centering
  \caption{Root mean square error (RMSE) of the predicted shadow matte image compared to the ground-truth image on the RDD and Kligler datasets.}
  \label{tab:shadow_matte_rmse}
  \begin{tabular}{@{}lcc@{}}
    \toprule
    Dataset & RDD (512$\times$512) & Kligler (512$\times$512) \\
    \midrule
    RMSE $\downarrow$ & 0.025 & 0.045 \\
    \bottomrule
  \end{tabular}
\end{table}

\begin{table*}[t]
  \centering
  \small
  \caption{Results of various guidance types on the RDD and Kligler datasets.}
  \label{tab:matte_ablation}
  \begin{tabular}{@{}l ccc ccc@{}}
    \toprule
    \multirow{2}{*}{Guidance Type} 
    & \multicolumn{3}{c}{RDD (512$\times$512)} 
    & \multicolumn{3}{c}{Kligler (512$\times$512)} \\
    \cmidrule(lr){2-4} \cmidrule(lr){5-7}
    & PSNR $\uparrow$ & SSIM $\uparrow$ & RMSE $\downarrow$ 
    & PSNR $\uparrow$ & SSIM $\uparrow$ & RMSE $\downarrow$ \\
    \midrule
    Ours w/o Guidance & 32.6 & 0.96 & 6.55 & 26.68 & \textbf{0.94} & 13.87 \\
    Ours w/ Binary Mask & 32.73 & \textbf{0.97} & 6.75 & 27.17 & \textbf{0.94} & 12.81 \\
    Ours & \textbf{33.78} & \textbf{0.97} & \textbf{5.78} & \textbf{28.74} & \textbf{0.94} & \textbf{11.19} \\
    \bottomrule
  \end{tabular}
\end{table*}

We evaluated the performance of the shadow matte generator in the proposed framework. \cref{tab:shadow_matte_rmse} reports the RMSE values for the predicted shadow matte against the ground truth image on the RDD and Kligler datasets.  The RMSE values are remarkably low (0.025 for RDD and 0.045 for Kligler), indicating that the shadow matte generator precisely and accurately captures luminance differences between shadowed and shadow-free regions. Such low errors suggest that the predicted mattes closely approximate the ground truth, which is essential for providing reliable guidance in the shadow removal process.

\begin{figure}[h]
    \centering
    
    \begin{subfigure}[b]{0.31\columnwidth}
        \includegraphics[width=\textwidth]{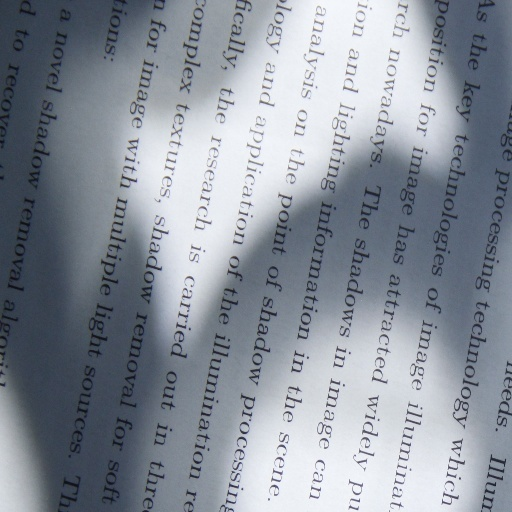}
        \vspace{1mm}
        \includegraphics[width=\textwidth]{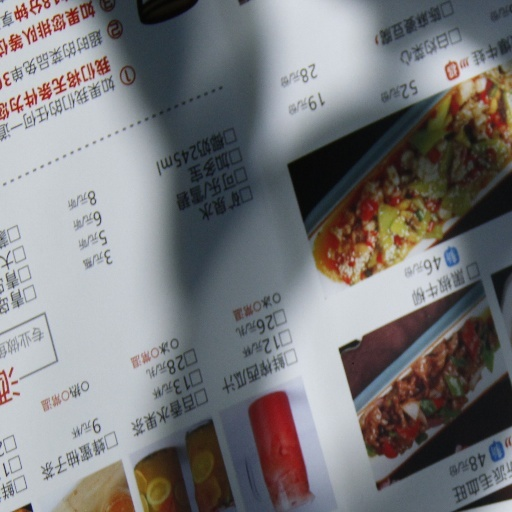}
        \caption{Input}
        \label{fig:matte_input}
    \end{subfigure}
    \hfill
    \begin{subfigure}[b]{0.31\columnwidth}
        \includegraphics[width=\textwidth]{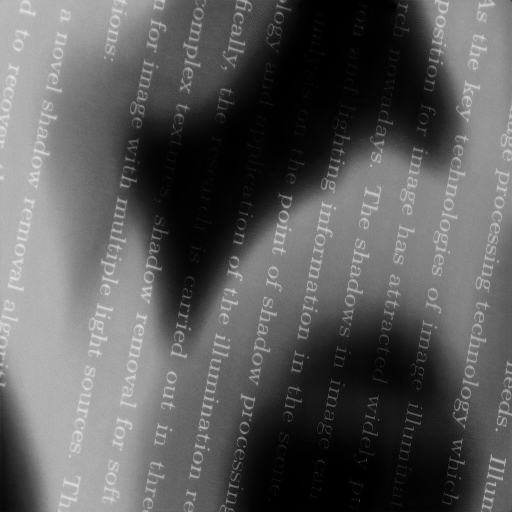}
        \vspace{1mm}
        \includegraphics[width=\textwidth]{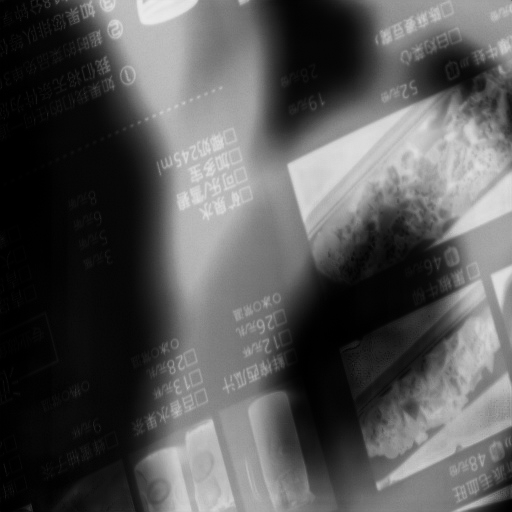}
        \caption{Pred}
        \label{fig:matte_pred}
    \end{subfigure}
    \hfill
    \begin{subfigure}[b]{0.31\columnwidth}
        \includegraphics[width=\textwidth]{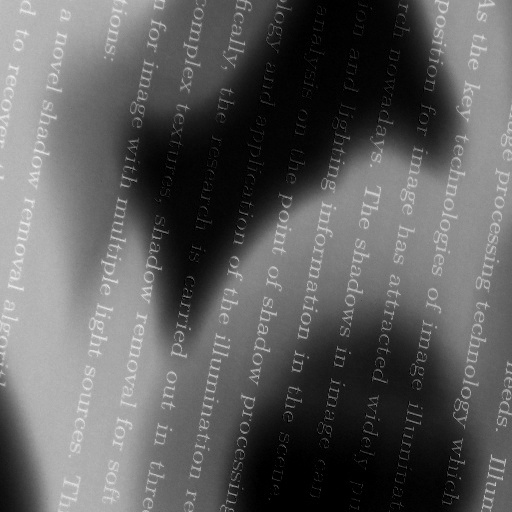}
        \vspace{1mm}
        \includegraphics[width=\textwidth]{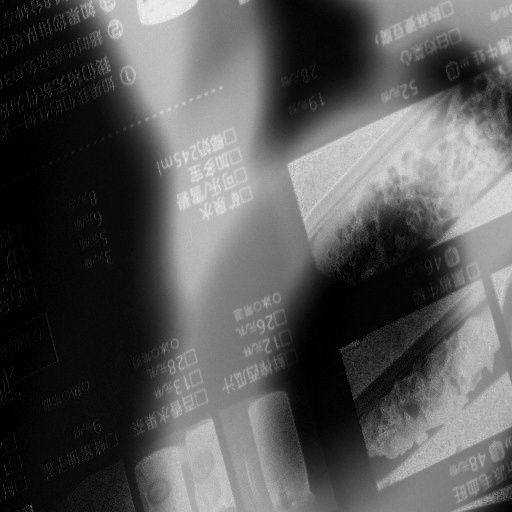}
        \caption{Target}
        \label{fig:matte_target}
    \end{subfigure}
    
    \caption{Qualitative results of shadow matte prediction. The (b) predicted matte closely matches the (c) target, capturing shadow regions and their intensities while preserving fine details.}
    \label{fig:matte_results}
\end{figure}

\cref{fig:matte_results} presents qualitative examples of the predicted shadow matte on the RDD dataset, demonstrating that it captures shadow regions and their intensities and preserves fine details (e.g., text structures).  These results confirm that the shadow matte generator produces precise and informative guidance, contributing significantly to the overall performance of the proposed framework.

\begin{figure}[b]
    \centering
    
    % tabular 컬럼 사이의 기본 간격을 3pt로 설정
    \setlength{\tabcolsep}{3pt}

    %--- 두 컬럼 모두 'm' (수직 중앙) 정렬을 사용합니다 ---
    \begin{tabular}{m{0.15\columnwidth} m{0.82\columnwidth}}
        
        % --- 1. 윗줄 (Input) ---
        \centering \small (a) Input &
        \includegraphics[width=0.32\linewidth]{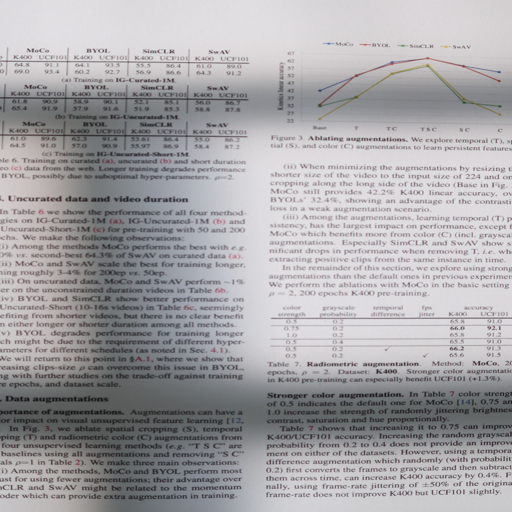}
        \hfill
        \includegraphics[width=0.32\linewidth]{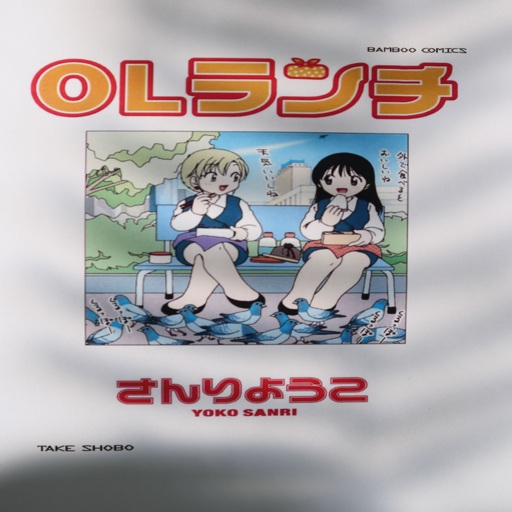}
        \hfill
        \includegraphics[width=0.32\linewidth]{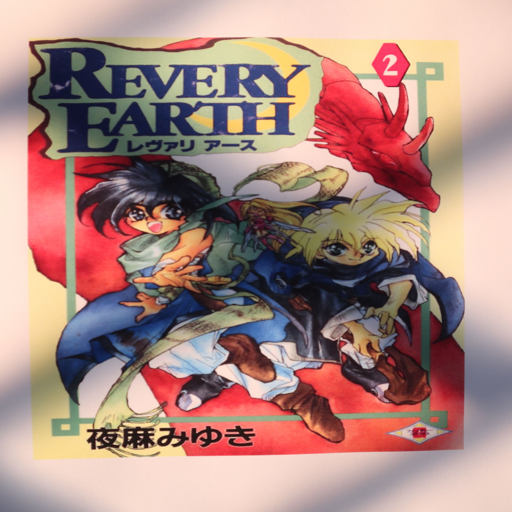} \\
        
        % booktabs의 \addlinespace로 두 행 사이에 수직 간격을 줍니다.
        \addlinespace[3pt]
        
        % --- 2. 아랫줄 (Pred) ---
        \centering \small (b) Pred &
        \includegraphics[width=0.32\linewidth]{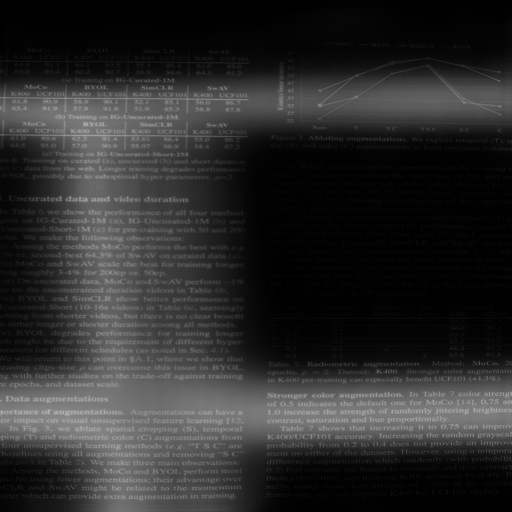}
        \hfill
        \includegraphics[width=0.32\linewidth]{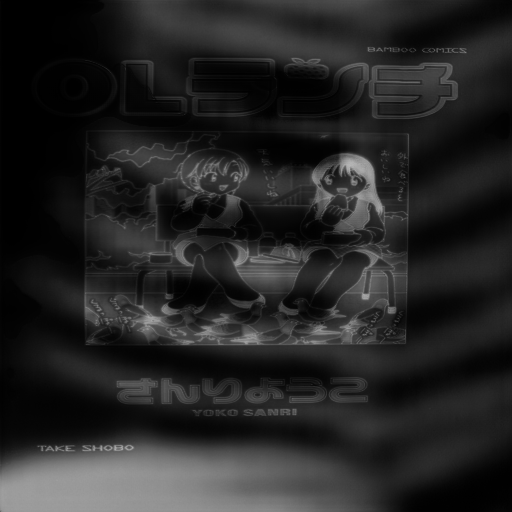}
        \hfill
        \includegraphics[width=0.32\linewidth]{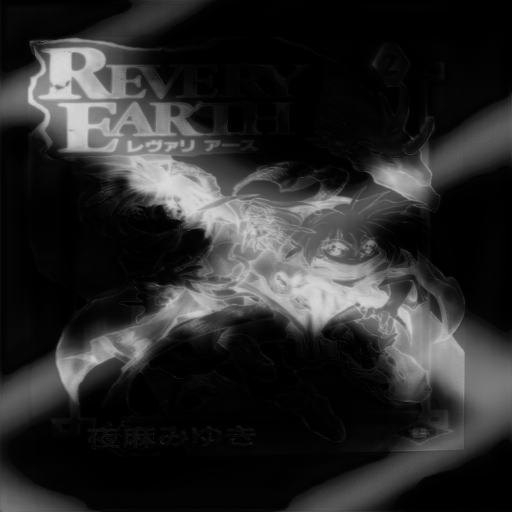}
        
    \end{tabular}
    
    \caption{Qualitative results of our shadow matte generator on the unseen SD7K dataset, showing (a) input images and (b) the corresponding predicted mattes. The generator was trained on the RDD dataset, demonstrating strong generalization to diverse, unseen document types.}
    \label{fig:sd7k_results}
\end{figure}

\begin{figure*}[t]
    \centering
    
    \begin{subfigure}[b]{0.23\textwidth}
        \includegraphics[width=\textwidth]{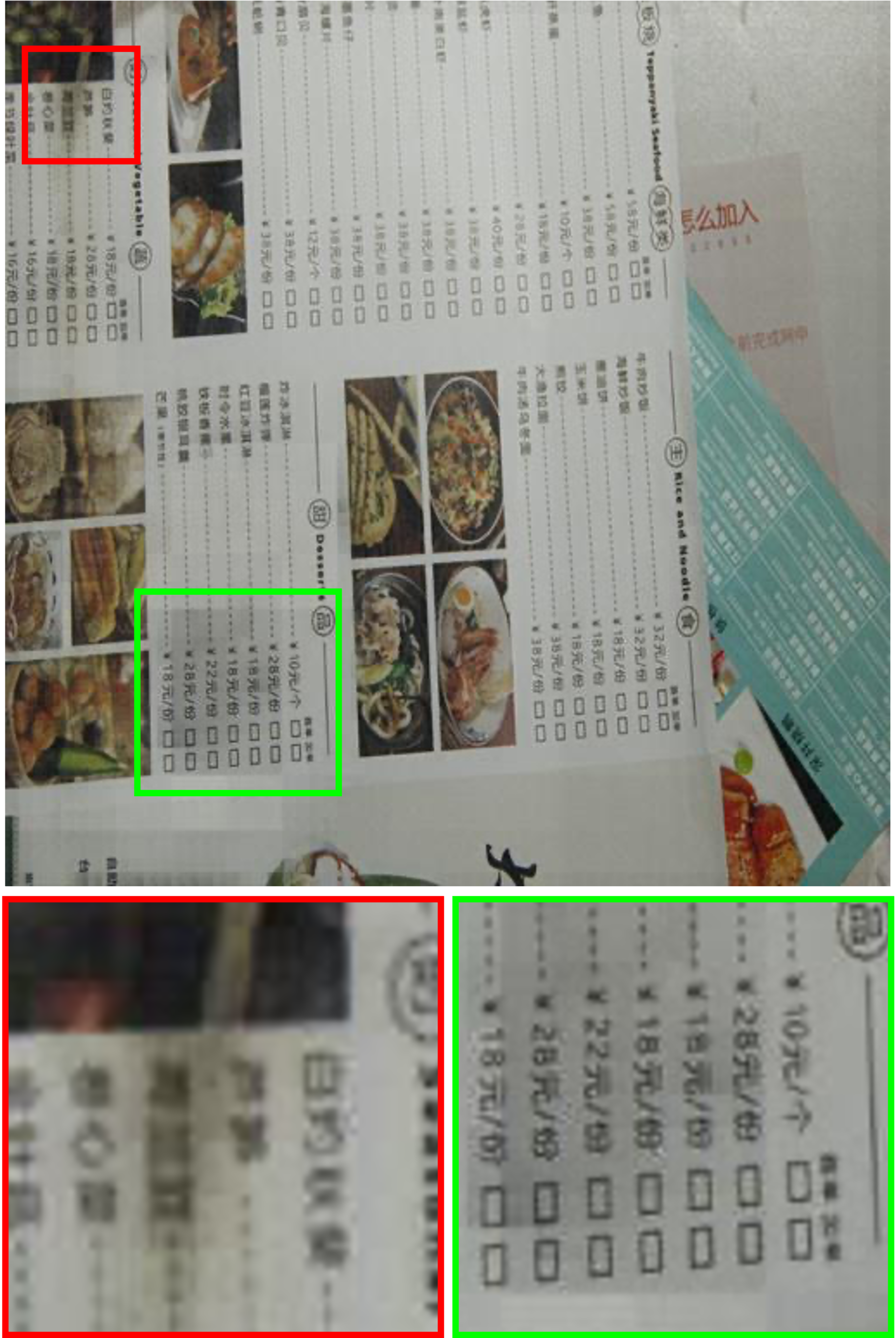}
        \caption{Ours w/o HFAM}
        \label{fig:hfam_no_1}
    \end{subfigure}
    \hfill
    \begin{subfigure}[b]{0.23\textwidth}
        \includegraphics[width=\textwidth]{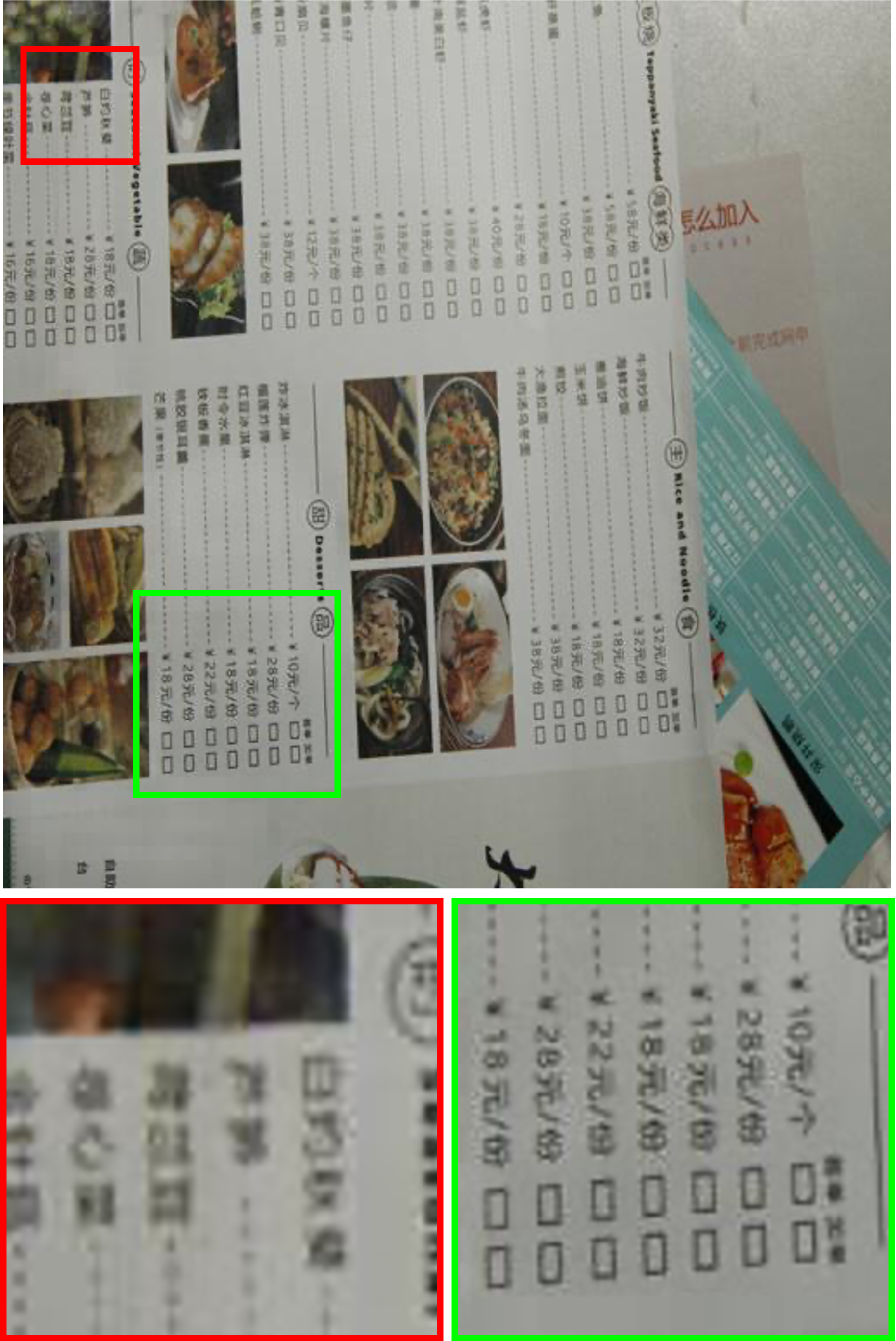}
        \caption{Ours (Full)}
        \label{fig:hfam_yes_1}
    \end{subfigure}
    \hfill % \hfill은 subfigure 사이에 위치해야 합니다.
    \begin{subfigure}[b]{0.23\textwidth}
        \includegraphics[width=\textwidth]{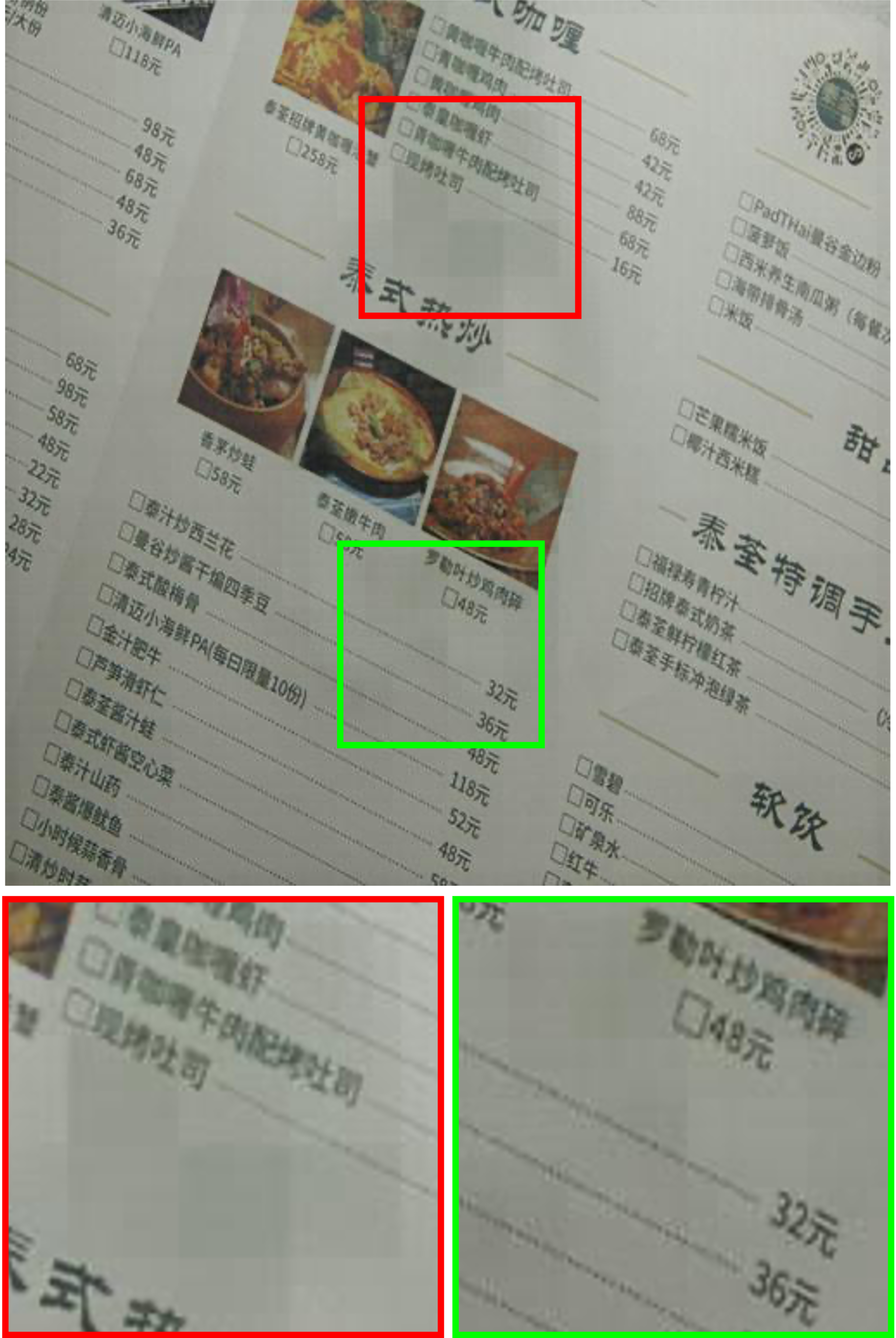}

        \setcounter{subfigure}{0} 
        
        \caption{Ours w/o HFAM}
        \label{fig:hfam_no_2}
    \end{subfigure}
    \hfill
    \begin{subfigure}[b]{0.23\textwidth}
        \includegraphics[width=\textwidth]{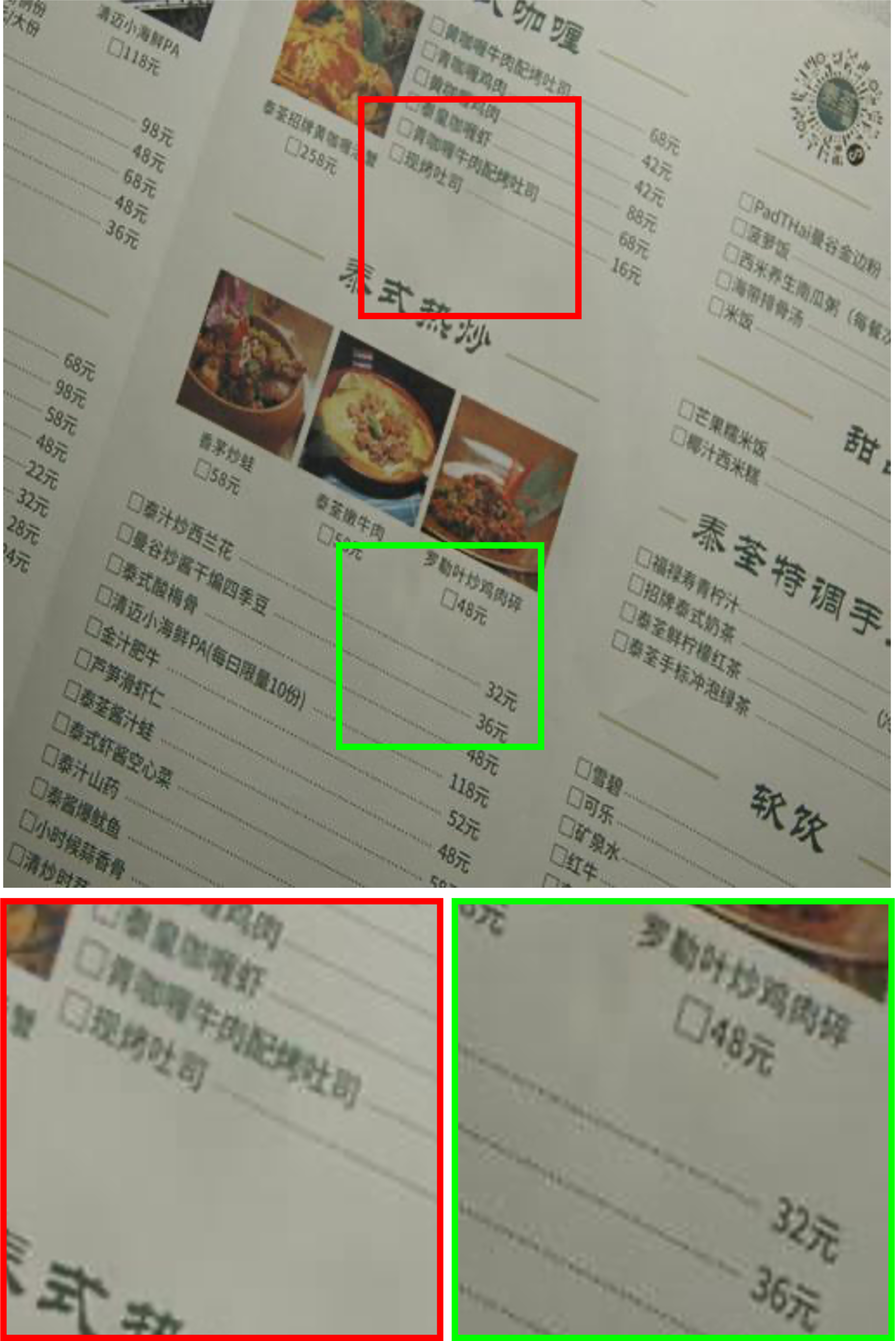}
        \caption{Ours (Full)}
        \label{fig:hfam_yes_2}
    \end{subfigure}

    \caption{Visual comparison for the ablation study on the high-frequency amplification module (HFAM). Qualitative results demonstrating the effect of the HFAM. (b) Full model with improved text sharpness, reduced shadow artifacts, and fewer block artifacts compared with (a) the model without HFAM.}
    \label{fig:hfam_comparison}
\end{figure*}

Furthermore, to evaluate the generalization performance and address the potential risk of the shadow matte generator overfitting to the training datasets, we conducted an additional qualitative evaluation on the recent, unseen SD7K dataset \cite{li2023high}. As shown in \cref{fig:sd7k_results}, our proposed generator accurately captures the shape and intensity of shadow regions even on images with diverse characteristics (e.g., technical documents, comic book covers) that differ from the training data. This suggests the model learned intrinsic shadow properties beyond the training data, confirming it as a robust and reliable pre-processing module for diverse real-world document images.

\subsection{Ablation Study}

\begin{table*}[t]
  \centering
  \small
  \caption{Ablation study results evaluating the influence of the high-frequency amplification module (HFAM) on the RDD and Kligler datasets.}
  \label{tab:hfam_ablation}
  \begin{tabular}{@{}l ccc ccc@{}}
    \toprule
    \multirow{2}{*}{Module} 
    & \multicolumn{3}{c}{RDD (512$\times$512)} 
    & \multicolumn{3}{c}{Kligler (512$\times$512)} \\
    \cmidrule(lr){2-4} \cmidrule(lr){5-7}
    & PSNR $\uparrow$ & SSIM $\uparrow$ & RMSE $\downarrow$ 
    & PSNR $\uparrow$ & SSIM $\uparrow$ & RMSE $\downarrow$ \\
    \midrule
    Ours w/o HFAM & 30.25 & 0.96 & 8.32 & 26.89 & 0.94 & 14.76 \\
    Ours & \textbf{33.78} & \textbf{0.97} & \textbf{5.78} & \textbf{28.74} & \textbf{0.94} & \textbf{11.19} \\
    \bottomrule
  \end{tabular}
\end{table*}

\paragraph{Effect of Shadow Matte Guidance} The ablation experiments compare the following three configurations to assess the contribution of the shadow matte as guidance: the proposed model without guidance, with a conventional binary shadow mask, and with the proposed shadow matte. \cref{tab:matte_ablation} summarizes the ablation results on the RDD and Kligler datasets.

For both datasets, the model with shadow matte guidance performs best across all evaluation metrics. The configuration without any guidance performs worst, especially on the Kligler dataset. On RDD, the binary mask improves PSNR and SSIM over the no-guidance baseline, but slightly worsens RMSE. Compared to Kligler, RDD shows a larger variation in shadow brightness and contains many faint shadows. This observation is consistent with recent studies on soft-shadow modeling, which have shown that hard binary masks cannot accurately capture soft, gradually brightening shadows \cite{wang2025softshadow, huang2025image}. Unlike binary masks, the shadow matte guidance captures subtle variations in shadow intensity and smooth transitions, facilitating better network localization of shadow regions and fine document detail preservation. This finding highlights the importance of applying fine-grained shadow information to achieve high-quality restoration in complex document scenarios.

\paragraph{Effect of HFAM} The effectiveness of the HFAM was evaluated via ablation experiments comparing the model with and without the HFAM.~\cref{tab:hfam_ablation} presents the results of the experiments on the RDD and Kligler datasets.

Incorporating HFAM consistently improves performance, with PSNR gains of 3.53 dB and 1.85 dB and RMSE reductions of 2.54 and 3.57 on the RDD and Kligler datasets, respectively. SSIM also increases from 0.96 to 0.97 on RDD.

These results confirm that the HFAM amplifies high-frequency details (e.g., text boundaries and fine structures), which are essential in document shadow removal tasks. By dynamically enhancing high-frequency regions, the HFAM helps the network retain sharper details and generate more accurate restorations, contributing to the quantitative gains and perceptual quality.

For further validation,~\cref{fig:hfam_comparison} visually compares these quantitative improvements. The results demonstrate that incorporating the HFAM significantly enhances restoration quality, particularly in shadow-affected or text-dense regions. Compared to the model without the HFAM, the full model produces noticeably sharper and more readable text. Blurred or smeared characters are refined, and background textures are more clearly restored, which is critical for document readability and analysis.

In addition, the model without the HFAM tends to produce block artifacts, a widespread problem arising from patch-wise processing. These artifacts are substantially reduced using the HFAM, suggesting that it captures and reconstructs global-local contextual information while preserving high-frequency components.
\section{Conclusion}
\label{sec:conclusion}

This paper proposes MatteViT, a novel framework for document shadow removal that applies shadow matte guidance and frequency-aware enhancement techniques, such as the HFAM and frequency-sensitive loss functions. By incorporating the HFAM, the proposed method preserves fine details and improves shadow removal quality. We also created a new shadow matte dataset and introduced a continuous shadow matte representation to enable shadow matte guidance. The extensive experiments demonstrate that the proposed approach achieves state-of-the-art performance across multiple metrics on the benchmark datasets. Furthermore, MatteViT improves OCR accuracy by better preserving text details, underlining its practical value in real-world document analysis. This framework offers an auspicious direction for robust document image enhancement and can be further extended to practical document digitization scenarios, such as mobile scanning applications and digital archiving systems.

\paragraph{Limitations} Our method may still miss some shadows on highly curved pages, such as those in folded books.
{
    \small
    \bibliographystyle{ieeenat_fullname}
    \bibliography{main}

@String(ICCV= {Int. Conf. Comput. Vis.})

@String(ICPR = {Int. Conf. Pattern Recog.})

@String(ICASSP=	{ICASSP})

@String(ICIP = {IEEE Int. Conf. Image Process.})

@String(AAAI = {AAAI})

@String(ICCV  = {ICCV})

@String(ICPR  = {ICPR})

@String(ICIP  = {ICIP})

@inproceedings{li2023high,
  title={High-resolution document shadow removal via a large-scale real-world dataset and a frequency-aware shadow erasing net},
  author={Li, Zinuo and Chen, Xuhang and Pun, Chi-Man and Cun, Xiaodong},
  booktitle={2023 IEEE/CVF International Conference on Computer Vision (ICCV)},
  pages={12415--12424},
  year={2023},
  organization={IEEE}
}

@inproceedings{zhang2023document,
  title={Document image shadow removal guided by color-aware background},
  author={Zhang, Ling and He, Yinghao and Zhang, Qing and Liu, Zheng and Zhang, Xiaolong and Xiao, Chunxia},
  booktitle={Proceedings of the IEEE/CVF Conference on Computer Vision and Pattern Recognition},
  pages={1818--1827},
  year={2023}
}

@inproceedings{wang2022udoc,
  title={UDoc-GAN: Unpaired document illumination correction with background light prior},
  author={Wang, Yonghui and Zhou, Wengang and Lu, Zhenbo and Li, Houqiang},
  booktitle={Proceedings of the 30th ACM International Conference on Multimedia},
  pages={5074--5082},
  year={2022}
}

@article{Anvari2021arXiv,
  author  = {Anvari Zahra and Vassilis Athitsos},
  title = {A survey on deep learning based document image enhancement},
  journal = {arXiv preprint arXiv:2112.02719},
  year    = {2021}
}

@inproceedings{pei2023doc,
  title={Doc-Former: A transformer-based document shadow denoising network},
  author={Pei, Shengchang and Liu, Jun and Yi, Niannian and Zhang, Yun and Liu, Zhengtao and Chen, Zengyan},
  booktitle={Proceedings of the 2023 6th International Conference on Robot Systems and Applications},
  pages={139--143},
  year={2023}
}

@article{feng2021doctr,
  title={DocTr: Document image transformer for geometric unwarping and illumination correction},
  author={Feng, Hao and Wang, Yuechen and Zhou, Wengang and Deng, Jiajun and Li, Houqiang},
  journal={arXiv preprint arXiv:2110.12942},
  year={2021}
}

@inproceedings{souibgui2022docentr,
  title={DocEnTr: An end-to-end document image enhancement transformer},
  author={Souibgui, Mohamed Ali and Biswas, Sanket and Jemni, Sana Khamekhem and Kessentini, Yousri and Forn{\'e}s, Alicia and Llad{\'o}s, Josep and Pal, Umapada},
  booktitle={2022 26th International Conference on Pattern Recognition (ICPR)},
  pages={1699--1705},
  year={2022},
  organization={IEEE}
}

@inproceedings{zhou2024docdeshadower,
  title={DocDeshadower: Frequency-aware transformer for document shadow removal},
  author={Zhou, Ziyang and Lei, Yingtie and Chen, Xuhang and Luo, Shenghong and Zhang, Wenjun and Pun, Chi-Man and Wang, Zhen},
  booktitle={2024 IEEE International Conference on Systems, Man, and Cybernetics (SMC)},
  pages={2468--2473},
  year={2024},
  organization={IEEE}
}

@inproceedings{chen2024shadocformer,
  title={ShaDocFormer: A shadow-attentive threshold detector with cascaded fusion refiner for document shadow removal},
  author={Chen, Weiwen and Lei, Yingtie and Luo, Shenghong and Zhou, Ziyang and Li, Mingxian and Pun, Chi-Man},
  booktitle={2024 International Joint Conference on Neural Networks (IJCNN)},
  pages={1--8},
  year={2024},
  organization={IEEE}
}

@inproceedings{yang2023docdiff,
  title={DocDiff: Document enhancement via residual diffusion models},
  author={Yang, Zongyuan and Liu, Baolin and Xxiong, Yongping and Yi, Lan and Wu, Guibin and Tang, Xiaojun and Liu, Ziqi and Zhou, Junjie and Zhang, Xing},
  booktitle={Proceedings of the 31st ACM international conference on multimedia},
  pages={2795--2806},
  year={2023}
}

@inproceedings{fu2019cascaded,
  title={Cascaded detail-preserving networks for super-resolution of document images},
  author={Fu, Zhichao and Kong, Yu and Zheng, Yingbin and Ye, Hao and Hu, Wenxin and Yang, Jing and He, Liang},
  booktitle={2019 International Conference on Document Analysis and Recognition (ICDAR)},
  pages={240--245},
  year={2019},
  organization={IEEE}
}

@inproceedings{chang2023tsrformer,
  title={TSRFormer: Transformer based two-stage refinement for single image shadow removal},
  author={Chang, Hua-En and Hsieh, Chia-Hsuan and Yang, Hao-Hsiang and Chen, I and Chen, Yi-Chung and Chiang, Yuan-Chun and Huang, Zhi-Kai and Chen, Wei-Ting and Kuo, Sy-Yen and others},
  booktitle={Proceedings of the IEEE/CVF Conference on Computer Vision and Pattern Recognition},
  pages={1436--1446},
  year={2023}
}

@inproceedings{le2020shadow,
  title={From shadow segmentation to shadow removal},
  author={Le, Hieu and Samaras, Dimitris},
  booktitle={European Conference on Computer Vision},
  pages={264--281},
  year={2020},
  organization={Springer}
}

@inproceedings{guo2023shadowdiffusion,
  title={ShadowDiffusion: When degradation prior meets diffusion model for shadow removal},
  author={Guo, Lanqing and Wang, Chong and Yang, Wenhan and Huang, Siyu and Wang, Yufei and Pfister, Hanspeter and Wen, Bihan},
  booktitle={Proceedings of the IEEE/CVF Conference on Computer Vision and Pattern Recognition},
  pages={14049--14058},
  year={2023}
}

@inproceedings{qu2017deshadownet,
  title={DeshadowNet: A multi-context embedding deep network for shadow removal},
  author={Qu, Liangqiong and Tian, Jiandong and He, Shengfeng and Tang, Yandong and Lau, Rynson WH},
  booktitle={Proceedings of the IEEE Conference on Computer Vision and Pattern Recognition},
  pages={4067--4075},
  year={2017}
}

@article{liu2021shadow,
  title={Shadow removal by a lightness-guided network with training on unpaired data},
  author={Liu, Zhihao and Yin, Hui and Mi, Yang and Pu, Mengyang and Wang, Song},
  journal={IEEE Transactions on Image Processing},
  volume={30},
  pages={1853--1865},
  year={2021},
  publisher={IEEE}
}

@article{liu2023decoupled,
  title={A decoupled multi-task network for shadow removal},
  author={Liu, Jiawei and Wang, Qiang and Fan, Huijie and Li, Wentao and Qu, Liangqiong and Tang, Yandong},
  journal={IEEE Transactions on Multimedia},
  volume={25},
  pages={9449--9463},
  year={2023},
  publisher={IEEE}
}

@inproceedings{hu2019mask,
  title={Mask-ShadowGAN: Learning to remove shadows from unpaired data},
  author={Hu, Xiaowei and Jiang, Yitong and Fu, Chi-Wing and Heng, Pheng-Ann},
  booktitle={Proceedings of the IEEE/CVF International Conference on Computer Vision},
  pages={2472--2481},
  year={2019}
}

@inproceedings{lin2020bedsr,
  title={BEDSR-Net: A deep shadow removal network from a single document image},
  author={Lin, Yun-Hsuan and Chen, Wen-Chin and Chuang, Yung-Yu},
  booktitle={Proceedings of the IEEE/CVF conference on computer vision and pattern recognition},
  pages={12905--12914},
  year={2020}
}

@inproceedings{guo2023shadowformer,
  title={ShadowFormer: Global context helps shadow removal},
  author={Guo, Lanqing and Huang, Siyu and Liu, Ding and Cheng, Hao and Wen, Bihan},
  booktitle={Proceedings of the AAAI conference on artificial intelligence},
  volume={37},
  number={1},
  pages={710--718},
  year={2023}
}

@article{dosovitskiy2020image,
  title={An image is worth 16x16 words: Transformers for image recognition at scale},
  author={Dosovitskiy, Alexey and Beyer, Lucas and Kolesnikov, Alexander and Weissenborn, Dirk and Zhai, Xiaohua and Unterthiner, Thomas and Dehghani, Mostafa and Minderer, Matthias and Heigold, Georg and Gelly, Sylvain and others},
  journal={arXiv preprint arXiv:2010.11929},
  year={2020}
}

@inproceedings{ronneberger2015u,
  title={U-Net: Convolutional networks for biomedical image segmentation},
  author={Ronneberger, Olaf and Fischer, Philipp and Brox, Thomas},
  booktitle={International Conference on Medical Image Computing and Computer-Assisted Intervention},
  pages={234--241},
  year={2015},
  organization={Springer}
}

@inproceedings{perez2018film,
  title={FiLM: Visual reasoning with a general conditioning layer},
  author={Perez, Ethan and Strub, Florian and De Vries, Harm and Dumoulin, Vincent and Courville, Aaron},
  booktitle={Proceedings of the AAAI conference on artificial intelligence},
  volume={32},
  number={1},
  year={2018}
}

@article{charbonnier1997deterministic,
  title={Deterministic edge-preserving regularization in computed imaging},
  author={Charbonnier, Pierre and Blanc-F{\'e}raud, Laure and Aubert, Gilles and Barlaud, Michel},
  journal={IEEE Transactions on Image Processing},
  volume={6},
  number={2},
  pages={298--311},
  year={1997},
  publisher={IEEE}
}

@inproceedings{shah2018iterative,
  title={An iterative approach for shadow removal in document images},
  author={Shah, Vatsal and Gandhi, Vineet},
  booktitle={2018 IEEE International Conference on Acoustics, Speech and Signal Processing (ICASSP)},
  pages={1892--1896},
  year={2018},
  organization={IEEE}
}

@inproceedings{kligler2018document,
  title={Document enhancement using visibility detection},
  author={Kligler, Netanel and Katz, Sagi and Tal, Ayellet},
  booktitle={Proceedings of the IEEE Conference on Computer Vision and Pattern Recognition},
  pages={2374--2382},
  year={2018}
}

@inproceedings{smith2007overview,
  title={An overview of the Tesseract OCR engine},
  author={Smith, Ray},
  booktitle={Ninth International Conference on Document Analysis and Recognition (ICDAR 2007)},
  volume={2},
  pages={629--633},
  year={2007},
  organization={IEEE}
}

@inproceedings{chen2023shadocnet,
  title={ShaDocNet: Learning spatial-aware tokens in transformer for document shadow removal},
  author={Chen, Xuhang and Cun, Xiaodong and Pun, Chi-Man and Wang, Shuqiang},
  booktitle={ICASSP 2023-2023 IEEE International Conference on Acoustics, Speech and Signal Processing (ICASSP)},
  pages={1--5},
  year={2023},
  organization={IEEE}
}

@article{wang2025comprehensive,
  title={A comprehensive survey on shadow removal from document images: datasets, methods, and opportunities},
  author={Wang, Bingshu and Li, Changping and Zou, Wenbin and Zhang, Yongjun and Chen, Xuhang and Chen, CL Philip},
  journal={Vicinagearth},
  volume={2},
  number={1},
  pages={1},
  year={2025},
  publisher={Springer}
}

@article{liu2025leveraging,
  title={Leveraging Contrast Information for Efficient Document Shadow Removal},
  author={Liu, Yifan and Huang, Jiancheng and Liu, Na and Yan, Mingfu and Huang, Yi and Chen, Shifeng},
  journal={arXiv preprint arXiv:2504.00385},
  year={2025}
}

@inproceedings{georgiadis2023lp,
  title={LP-IOANet: Efficient high resolution document shadow removal},
  author={Georgiadis, Konstantinos and Yucel, M Kerim and Skartados, Evangelos and Dimaridou, Valia and Drosou, Anastasios and Saa-Garriga, Albert and Manganelli, Bruno},
  booktitle={ICASSP 2023-2023 IEEE International Conference on Acoustics, Speech and Signal Processing (ICASSP)},
  pages={1--5},
  year={2023},
  organization={IEEE}
}

@inproceedings{lu2017shadow,
  title={A shadow removal method for tesseract text recognition},
  author={Lu, Huimin and Guo, Baofeng and Liu, Juntao and Yan, Xijun},
  booktitle={2017 10th International Congress on Image and Signal Processing, BioMedical Engineering and Informatics (CISP-BMEI)},
  pages={1--5},
  year={2017},
  organization={IEEE}
}

@inproceedings{jung2018water,
  title={Water-filling: An efficient algorithm for digitized document shadow removal},
  author={Jung, Seungjun and Hasan, Muhammad Abul and Kim, Changick},
  booktitle={Asian Conference on Computer Vision},
  pages={398--414},
  year={2018},
  organization={Springer}
}

@inproceedings{wang2019effective,
  title={An effective background estimation method for shadows removal of document images},
  author={Wang, Bingshu and Chen, CL Philip},
  booktitle={2019 IEEE International Conference on Image Processing (ICIP)},
  pages={3611--3615},
  year={2019},
  organization={IEEE}
}

@inproceedings{wang2020shadow,
  title={Shadow removal of text document images by estimating local and global background colors},
  author={Wang, Jian-Ren and Chuang, Yung-Yu},
  booktitle={ICASSP 2020-2020 IEEE International Conference on Acoustics, Speech and Signal Processing (ICASSP)},
  pages={1534--1538},
  year={2020},
  organization={IEEE}
}

@inproceedings{liu2023shadow,
  title={Shadow removal of text document images using background estimation and adaptive text enhancement},
  author={Liu, Wenjie and Wang, Bingshu and Zheng, Jiangbin and Wang, Wenmin},
  booktitle={ICASSP 2023-2023 IEEE International Conference on Acoustics, Speech and Signal Processing (ICASSP)},
  pages={1--5},
  year={2023},
  organization={IEEE}
}

@article{qi2020stc,
  title={STC-GAN: Spatio-temporally coupled generative adversarial networks for predictive scene parsing},
  author={Qi, Mengshi and Wang, Yunhong and Li, Annan and Luo, Jiebo},
  journal={IEEE Transactions on Image Processing},
  volume={29},
  pages={5420--5430},
  year={2020},
  publisher={IEEE}
}

@inproceedings{wang2025softshadow,
  title={SoftShadow: Leveraging Soft Masks for Penumbra-Aware Shadow Removal},
  author={Wang, Xinrui and Guo, Lanqing and Wang, Xiyu and Huang, Siyu and Wen, Bihan},
  booktitle={Proceedings of the Computer Vision and Pattern Recognition Conference},
  pages={23217--23226},
  year={2025}
}

@article{huang2025image,
  title={Image shadow removal via multi-scale deep retinex decomposition},
  author={Huang, Yan and Lu, Xinchang and Quan, Yuhui and Xu, Yong and Ji, Hui},
  journal={Pattern Recognition},
  volume={159},
  pages={111126},
  year={2025},
  publisher={Elsevier}
}
}

\end{document}